\documentclass{article}

\usepackage[preprint]{neurips_2026}

\usepackage[utf8]{inputenc} %
\usepackage[T1]{fontenc}    %
\usepackage{hyperref}       %
\usepackage{url}            %
\usepackage{booktabs}       %
\usepackage{amsfonts}       %
\usepackage{nicefrac}       %
\usepackage{microtype}      %
\usepackage{xcolor}         %
\usepackage{amsmath}
\usepackage{graphicx}
\usepackage{algorithm}
\usepackage{algpseudocode}
\usepackage{enumitem}
\usepackage{multirow}

\newtheorem{lemma}{Lemma}
\newtheorem{theorem}{Theorem}

\usepackage{placeins}

\newcommand{\eat}[1]{}

\title{Halt Fast! Early Stopping for Certified Robustness}

\author{%
  Andrew C. Cullen \\
  University of Melbourne \\
  \texttt{andrew.cullen@unimelb.edu.au} \\
  \And
   Paul Montague \\
   DST Group, Adelaide \\
   \And
  Benjamin I. P. Rubinstein \\
  University of Melbourne \\
}

\begin{document}

\maketitle

\begin{abstract}
Randomized Smoothing (RS) provides rigorous robustness guarantees for neural networks without architectural constraints, yet its adoption is limited by extreme computational costs. Standard RS requires tens of thousands of model evaluations per input and forces practitioners to commit to fixed sample sizes \emph{a priori}. In this work, we present a novel meta-learning framework for anytime-valid certified robustness that adaptively deploys computational resources. By using a lightweight meta-learner to predict image-specific priors for a sequential E-process, we achieve a 20-fold reduction in sample complexity compared to traditional methods while maintaining rigorous statistical guarantees. Beyond raw efficiency, we demonstrate how anytime-validity enables adaptively allocating compute based upon application-specific risk thresholds, a form of resource triage impossible under classic certification frameworks. That this is achievable while also providing similar certification performance demonstrates that our approach provides a pathway for real-time, safety-critical certification deployments.

\end{abstract}

\section{Introduction}

For all of their transformative utility, neural networks remain notoriously sensitive to oftentimes semantically meaningless modifications~\citep{szegedy2013intriguing}. These modifications, now known as adversarial examples~\citep{goodfellow2014explaining, madry2017towards}, have spurred research that has consistently demonstrated that model decision boundaries often lack the semantic alignment required for safety-critical deployments. 

While defenses have been proposed against these manipulations, they are fundamentally constrained within a technological arms race, where each new defense provides something new to attack~\citep{goodfellow2014explaining, cullen25a}. By contrast, Certified Robustness has emerged as a rigorous framework for mathematically guaranteeing that a model's prediction remains invariant within a defined neighborhood of an input~\citep{lecuyer2019certified, cohen2019certified, cullen2022double}. In the case of classification systems, this formulation is typically defined in terms of a radius $r$ such that for a model $F$ we can guarantee that $F(x) = F(x')$ for all $x'$ in the ball $B_{p}(x, r) = \{x' : \| x' - x\|_p \le r\}$.

Among certifications, Randomised Smoothing (RS) is unique in that it can be applied to any model without architectural modifications. However, this flexibility comes at a cost---to complete a certification, a sample must be passed through the model tens of thousands of times, with each copy being offset by a small perturbation drawn from the Normal distribution. This high sample complexity, which is required to control for the Type-I (false positive) errors render real-time applications of certifications nigh-on impossible, especially for models at scale.

Recent advances have sought to alleviate this overhead through sequential testing and early stopping. Most notably, the introduction of \textbf{E-values}~\citep{shafer2019game, ramdas2023game} and test martingales have allowed for anytime-valid certifications~\citep{voravcek2024treatment}. By framing certification as a super-martingale wealth process, sampling can be halted the moment sufficient evidence for robustness is accumulated without violating statistical safety. However, current E-value applications in robustness have primarily focused on binary hypothesis testing (e.g., asking if $r \geq c$ or not), reducing the certifier to a simple threshold-based classifier. This limitation robs the process of the discriminatory detail required to assess relative security across different samples. %

In this work, we argue that the primary utility of anytime-valid certifications is not necessarily their computational efficiency, but rather the ability to adaptively shape termination conditions around application-based workflows. In aid of this goal, our contributions are three-fold:
\begin{enumerate}[leftmargin=*, noitemsep, topsep=0pt]
    \item \textbf{Method of Mixtures for Continuous Hypotheses:} We extend E-value certifications to a mixture-based multiple-hypothesis approach to match traditional certification workflows.%
    \item \textbf{Sample-Adaptive Meta-Learning:} We introduce an optimized E-value formulation, where a meta-learner predicts the prior distribution to enhance efficiency. We employ a lightweight meta-learner to analyze initial model glimpses, utilizing a Bayesian Negative Log-Likelihood (NLL) objective to fit the distribution of the smoothed model's success rate for a given input. 
    \item \textbf{Adversarial Exits:} We introduce task-adaptive termination conditions, that allow for early-termination based upon prespecified domain tasks, producing highly efficient certifications and rejections in a manner that optimizes global compute. 
\end{enumerate}
Through experimental validation, we demonstrate that our approach can construct certifications significantly faster than fixed-sample methods---demonstrating that viable certifications can be constructed in less than $500$ samples, a $20$-fold decrease over prior certification workflows. Perhaps more importantly, our innovations allow for application- and sample-specific exit conditions, which are a crucial innovation for helping certifications transition from a numerically-costly concept to a viable framework for producing real-world security.

\subsection{Motivating Cases}

To underscore the utility of the methods introduced within this paper, we highlight three scenarios where anytime-valid certifications are well motivated. The first is the most natural: \textbf{computational efficiency}. By halting as soon as a target precision is reached, we remove a primary barrier to real-world deployment. The second is \textbf{resource triage}: in large-scale systems, certificates can be used to route inputs into different verification pathways based on their robustness. In such cases, proving that a sample falls within a specific risk bucket is more critical than its exact radius. Finally, we also suggest that \textbf{streaming contexts} (e.g., autonomous driving) could also be an application of this approach, where prior information from preceding frames can be used to initialize E-value certifications, allowing for faster convergence in temporally evolving environments.

\section{Related Work}\label{sec:related}

RS has evolved from its Differential Privacy foundations~\citep{lecuyer2019certified, dwork2006calibrating} to the current state-of-the-art based on the Neyman-Pearson lemma~\citep{cohen2019certified}. At their core, all RS-based certifications transform a base model $f$ into a smoothed counterpart $g$ with provable $\ell_p$ margin guarantees. As established by \citet{cohen2019certified}, for a noise level $\sigma$, the certified radius $r$ is a function of the success probability $p_A = \mathbb{P}_{\epsilon \sim \mathcal{N}(0, \sigma^2 I)}[f(x+\epsilon) = c_A]$ of the most likely class $c_A$
\begin{equation}
    r = \sigma \Phi^{-1}(p_A)\enspace.
\end{equation}
In practice, certifications are constructed through an independent two-phase approach, where Phase I employs an initial batch to establish the target class, before then estimating $p_A$ through Monte-Carlo sampling in Phase II. To control the risk of false certification, standard approaches use the Clopper-Pearson interval~\citep{clopper1934use} to obtain a high-probability lower bound $\underline{p_A}$.%

The need for tight lower bounds is the primary driver of RS's high computational cost. High-variance inputs may require tens of thousands of samples to produce meaningful certificates. Crucially, the peaking problem~\citep{johari2017peeking} invalidates the Clopper-Pearson bounds on the Type-I error rate if a practitioner monitors the empirical mean and stops early.
Consequently, $N$ must be fixed \emph{a priori}, leading to massive over-sampling for easy inputs and total failure to certify for marginal ones.

In response, sequential testing has emerged as a pathway to efficient robustness.  To achieve this, the significance budget $\alpha$ is partitioned across multiple distinct stopping points $\{n_1, n_2, \ldots, n_k\}$ by way of the Bonferroni correction or alpha-spending functions \citep{horvath2022boosting}. While these approaches allow for early stopping, they still retain the fundamental drawback of the frequentist framework: after each $n_{i}$, if the model fails to certify, then the model samples further to $n_{i+1}$. However, the correction for examining multiple stopping points requires $\alpha$ to be scaled by the number of potential comparisons (which also must be set \emph{a priori}), increasing the number of samples required to certify to a given level. As such, these early-stopping frameworks potentially require significantly more net samples to be evaluated in order to certify a sample than a more naive implementation of \citet{cohen2019certified}. While additional efficiency can be found in \emph{a priori} estimation of appropriate sample counts~\citep{chen2022input}, %
these approaches are still inherently conservative and expensive.

E-values provide a natural response to circumvent these limitations, in that they are anytime-valid and immune to the peeking problem. This allows certifications to be constructed in a manner that allows for early-stopping, based upon arbitrary criteria. The original progenitor of this approach, \citet{voravcek2024treatment} primarily considered how this could be applied to binary robustness hypotheses (e.g., $r \ge r_0$).

In this work we leverage the \textbf{Method of Mixtures}~\citep{waudby2024estimating, grunwald2020safe} to support continuous radius estimation. Notably, prior mixture-based approaches rely on the Krichevsky–Trofimov (KT) estimator~\citep{krichevsky1981performance} which is optimized for \emph{arbitrary} sequences. However, we contend that RS sequences are not arbitrary; they are tied to specific input manifolds. This discrepancy is the theoretical foundation for our meta-learning framework, which learns to parameterize bespoke priors that sufficiently accelerate anytime-validity.%

For narrative clarity further discussions of certifications can be found in  Appendix~\ref{sec:extended_related}. %

\section{Anytime-Valid Radius Certification}\label{sec:anytime}

To evaluate the robustness of a model prediction of a class $c_A$ at a point $x$, we consider the success probability $p = P(f(x+\epsilon) = c_A)$ of the smoothed classifier (we stress that we presuppose knowledge of $c_A$ for mathematical convenience, and that the class is estimated appropriately in our algorithm). Assume that we observe an infinite sequence of i.i.d. Bernoulli trials $X_1, X_2, \ldots$, where each $X_i = \mathbb{I}[f(x+\epsilon_i) = c_A]$ denotes whether the $i$-th perturbation of $x$ with noise $\epsilon_i \sim \mathcal{N}(0, \sigma^2 I)$ aligns with the target class. To construct a certification, we must be able to test the null hypothesis $H_0: p \le p_0$ for any threshold $p_0 \in [0, 1]$ to construct an anytime-valid lower confidence bound on $p$.

\paragraph{Test Martingales and E-values}
To achieve this, we leverage the \emph{E-value}, a non-negative random variable $E$ such that $\mathbb{E}_{H_0}[E] \le 1$~\citep{vovk2021values, shafer2019language}. In our Bernoulli setting, for a point null hypothesis $H_0: p=p_0$ and a point alternative hypothesis $H_1: p=q$, the appropriate E-value is the likelihood ratio~\citep{waudby2024estimating} 
\begin{equation}
    E_i = \frac{q^{X_i}(1-q)^{1-X_i}}{p_0^{X_i}(1-p_0)^{1-X_i}}\enspace.
\end{equation}

To verify the significance of an accumulated E-value, we employ a process inspired by betting games~\citep{shafer2019language, vovk2021values}. Consider the wealth process $W_t(p_0) = \prod_{i=1}^t E_i$, that is the accumulation of E-values over $t$ samples. Then if $h_t = \sum_{i=1}^t X_i$ is the number of successes in the first $t$ trials, the total wealth must be
\begin{equation}
    W_{t}(p_0) = \frac{q^{h_t} (1-q)^{t-h_t}}{p_{0}^{h_t} (1 - p_0)^{t-h_t}}\enspace.
\end{equation}
This expression represents the ratio of the likelihood of observing the sequence under the alternative hypothesis versus the null hypothesis. %

The process $(W_t(p_0))_{t \ge 1}$ is a non-negative martingale with $\mathbb{E}[W_t] = 1$ under $p=p_0$. By Ville's inequality~\citep{ville1939etude, doob1940regularity}, the set of all $p_0$ for which the wealth has not yet crossed the rejection threshold $1/\alpha$ forms the confidence interval $C_{t} = \left\{ p_0 \in [0,1] : \max_{\tau \le t} W_{\tau}(p_0) < \frac{1}{\alpha} \right\}$ 
where the Lower Confidence Bound (LCB) is $\underline{p_{t}} = \inf C_{t}$.%

\begin{theorem}[Soundness]\label{thm:soundness}
For a target class $c_A$, significance level $\alpha \in (0, 1)$, and prior mixture $Q$, the lower confidence bound $\underline{p_t} = \inf \{ p_0 : \max_{\tau \leq t} \overline{W}_{\tau}(p_0) < 1/\alpha \}$ satisfies $P(\exists t : p < \underline{p_t}) \le \alpha$. %
\end{theorem}

This result follows from the \textit{test martingale inversion} principle~\citep{howard2021time, ramdas2023game}. For any fixed $p_0 \in [0,1]$,      the mixture wealth process $\bar{W}_t(p_0)$ is a non-negative martingale with $\bar{W}_0 = 1$ under the point null hypothesis $H_0: p = p_0$. By Ville's inequality, the      probability that the wealth ever exceeds $1/\alpha$ is bounded by $\alpha$. Since the confidence set $C_t$ is constructed by inverting this test, and by utilizing the running maximum of the wealth process, we ensure that the confidence bounds monotonically tighten over time, guaranteeing that the set contains the true success probability $p$ for all $t \geq 1$ with probability $1-\alpha$. Finally, because the mixture of E-values is generally log-convex with respect to $p_0$, the set of non-rejected hypotheses forms a contiguous interval, ensuring that the anytime-valid lower bound $\underline{p}_t = \inf C_t$  is well defined.

\paragraph{The Method of Mixtures}
In the context of certifications, we do not know \emph{a priori} what the alternative $q$ is, and yet knowing this on a sample-by-sample basis is crucially important for providing enough discriminatory information to support downstream applications. As such, we instead propose utilizing the \emph{Method of Mixtures}~\citep{grunwald2020safe, waudby2024estimating}, which tests against the set of hypotheses defined over a prior distribution $Q(q)$ of alternative hypotheses, by way of the mixture E-value
\begin{equation}
    \bar{W}_t(p_0) = \int_0^1 \frac{q^{h_t} (1-q)^{t-h_t}}{p_{0}^{h_t} (1 - p_0)^{t-h_t}} dQ(q) \qquad
    = \frac{m(h_t, t)}{p_0^{h_t} (1 - p_0)^{t-h_t}}\enspace,
\end{equation}
where $m(h, t) = \int_0^1 q^h (1-q)^{t-h} dQ(q)$ is the integrated likelihood.

\begin{lemma}[Mixture E-values]
For any probability measure $Q$ on $[0, 1]$, the mixture $\bar{W}_t(p_0)$ is a non-negative martingale under $H_0: p = p_0$, and thus an anytime-valid E-process.
\end{lemma}

In the case where $Q$ is a Beta distribution $B(\beta, \gamma)$, then the integrated likelihood admits the closed-form solution~\citep{lai1976confidence}
\begin{equation}
    m(h, t) = \frac{B(h + \beta, t - h + \gamma)}{B(\beta, \gamma)}\enspace,
\end{equation}
allowing wealth accumulation without the imprecision of numerical integration. 

While this solution is numerically beneficial, the convenience of using the Beta distribution to unimodal distributions, or very restricted bimodal options. Real-world success probabilities are rarely governed by a single mode; instead they may cluster into distinct groups under some contexts. To support flexible prior distributions in heterogeneously clustered data, we extend this approach to a mixture of $K$ Beta distributions, as any convex combination of E-values is also an E-value~\citep{vovk2021values, grunwald2020safe, waudby2024estimating}. 

\begin{lemma}[Convexity of E-values]
Let $E^{(1)}, E^{(2)}, \ldots, E^{(K)}$ be a collection of E-values for the same null hypothesis $H_0$. For any set of non-negative weights $w_k$ such that $\sum_{k=1}^K w_k = 1$, the weighted sum $E = \sum_{k=1}^K w_k E^{(k)}$ is also an E-value for $H_0$. 
\end{lemma}

The integrated likelihood under the $k$-th prior $m_k$ yields the total wealth (under $\sum w_k = 1$)
\begin{equation}
    \overline{W}_t(p_0) = \sum_{k=1}^K w_k \cdot \left( \frac{m_k(h_t, t)}{p_0^{h_t} (1 - p_0)^{t-h_t}} \right)\enspace,
\end{equation}

\section{Sample-Adaptive Meta-Learning}\label{sec:meta-learning}

The efficacy of the Mixture E-value process is fundamentally determined by how the probability mass $Q(q)$ is allocated across the space of alternatives to maximize wealth accumulation. Traditionally, this would be approached with the Krichevsky–Trofimov (KT) estimator~\citep{krichevsky1981performance}---equivalent to a $\text{Beta}(0.5, 0.5)$---as it will provably produce minimum regret for worst-case arbitrary sequences~\citep{xie2000asymptotic}.

However, we ask the question: \textbf{what happens if our inputs are not considered arbitrary?} Not only are they drawn from some distribution $\mathcal{D}$, we also have knowledge of the glimpse of $N_{sel}$ samples, taken through Phase I of the certification. If this information could be employed, then it would produce an information advantage relative to the naturally conservative KT prior~\citep{waudby2024estimating, grunwald2020safe}.

Drawing upon this intuition, we introduce a Sample-Adaptive Meta-Learner $\mathcal{M}_\theta$ to perform amortized Bayesian inference~\citep{gershman2014amortized}, predicting a bespoke mixture prior for each input. Our meta-learner leverages the Phase I information to parameterize a prior that can then inform the betting strategy for Phase II. Because these $N_{sel}$ samples are strictly discarded before wealth accumulation begins, the sequential process $W_t$ remains a predictable super-martingale~\citep{shafer2019game}, and the statistical guarantee of Theorem~\ref{thm:soundness} remains untainted by the prior's data-dependency~\citep{ramdas2023game}.

Our meta-learner ingests three distinct signals: the semantic context $\phi(x)$, representing the penultimate layer embedding of the base classifier, providing a high-dimensional representation of the image's difficulty and class; the raw softmax vector of the clean image $\mathbf{p}(x)$, representing the classifier confidence; and the empirical success rate $\hat{p}_{sel}$ observed over the glimpse of $N_{sel}$ samples.

Of these, the latter provides the most direct evidence of the true success probability $p$. Based upon our experiments, we extract scalar proxies from the classifier confidence $\mathbf{p}(x)$ vector so that it returns either the margin $p_{max} - p_{next}$  or the entropy, acting as proxies for the model's epistemic uncertainty. %
$\mathcal{M}_\theta$ can thus be trained through a Kelly Criterion based loss, which is equivalent to minimizing the Negative Log-Likelihood (NLL) of the binomial sequence $X_{1:N}$ under the predicted mixture of Beta distributions by
\begin{equation}\label{eqn:criterion_loss}
    \mathcal{L}(\theta) = - \mathbb{E}_{x \sim \mathcal{D}} \left[ \log \left( \sum_{k=1}^K w_k \frac{B(h_N + \beta_k, N - h_N + \gamma_k)}{B(\beta_k, \gamma_k)}    \right) \right] + \lambda_t \underbrace{\mathbb{E}_{x \sim \mathcal{D}}
     [ \sum_{k=1}^K \text{dist}(\hat{p}_{mle}, \mathcal{R}_k) ]}_{\text{Containment Penalty}}\enspace,
\end{equation}
where $(\beta_k, \gamma_k, w_k) = \mathcal{M}_\theta(\text{features})$, and $\lambda_t$ controls a regularizing penalty that minimizes the $\ell_1$ distance between $\hat{p}_{mle}$ and the nearest boundary of $\mathcal{R}_k$, where $\lambda_t$ starts at $10.0$ and linearly decays to $1.0$ over the first 80\% of training epochs. This formalism ensures that the expected growth rate of the wealth process is maximized~\citep{kelly1956new}, while penalizing over-confident predictions as the log-wealth approaches $-\infty$ when the prior mass is zero at the true $p$. Consequently, the meta-learner learns to output broader, more resilient priors for high-variance inputs while maintaining aggressive, concentrated bets for unambiguous samples so that it safely maximizes wealth accumulation. %
As the Phase I glimpse is inherently noisy, we augment the training process by sampling $M$ possible realizations of $\hat{p}_{sel}$ from the ground-truth bitstream by way of draws from a binomial distribution. This forces the meta-learner to learn a mapping $(\text{features}, \text{glimpse}) \to \text{Prior}$ that is robust to the variance of the initial sampling. To ensure the anytime-valid responsiveness of the mixture, we constrain the predicted Beta parameters to $\beta, \gamma \in [0.1, 500.0]$. The lower bound of $0.1$ allows the meta-learner to predict distributions that are sharper than the non-informative KT prior, while the upper bound of $500$ acts as a survival bias, preventing a single sampling anomaly triggering wealth bankruptcy. %

The meta-learner in our experiments takes the form of a 4-layer Residual MLP that maps these features to a $K$-component mixture prior. For each component $k$, the model predicts a weight $w_k$, Beta parameters $(\beta_k, \gamma_k)$, and---as will be discussed in Section~\ref{sec:truncated}---the boundaries of the support region $\mathcal{R}_{k}$.
Our formalism provides significant flexibility in how the meta-learner is constructed. Across our experiments, we consider frameworks where the number of E-values is varied, alongside%
\begin{itemize}[leftmargin=*, noitemsep, topsep=0pt]
    \item \textbf{Full Support}: Where the Beta distribution is defined over $p \in [0, 1]$.
    \item \textbf{Hybrid Support}: A fixed partition where components are pre-assigned to the robust $[0.5, 1.0]$ or non-robust $[0.0, 0.5]$ regions.%
    \item \textbf{Dynamic Support}: The model learns to focus on specific probability intervals, concentrating its betting resolution where it predicts the true $p$ will lie.
\end{itemize}
From this point on, we will employ the naming scheme \textbf{Meta-\{K\}-\{Support\}}, where K refers to the number of mixed distributions, and Support refers to one of the three convergence frameworks above.

\subsection{Truncated Betas}\label{sec:truncated}

To realize the partitioned support modes introduced in the preceding section, we formalize the prior $Q$ as a mixture of truncated Beta distributions. By restricting the probability mass of each mixture component $k$ to the specific (and potentially learned) interval $\mathcal{R}_{k} = [a_k, b_k]$, the meta-learner allocates probability mass to capture potential multi-modal distributions.%

A truncated Beta prior is equivalent to a standard Beta distribution that is rejection-sampled to lie within $[a_k, b_k]$, subject to normalization by the mass of the original distribution contained within the region $Z_k = I_{b_k}(\beta_k, \gamma_k) - I_{a_k}(\beta_k, \gamma_k)$,
where $I_x$ is the regularized incomplete Beta function (the CDF of the Beta distribution). Under this prior, the integrated likelihood $m_k(h_t, t; \mathcal{R}_k)$---the marginal likelihood of the data given the regional prior $Q_k$---admits the closed form%
\begin{equation}
    m_k(h_t, t; \mathcal{R}_k) = \underbrace{\frac{B(h_t + \beta_k, t - h_t + \gamma_k)}{B(\beta_k, \gamma_k)}}_{\text{Standard Beta Update}} \cdot \underbrace{\frac{I_{b_k}(h_t + \beta_k, t - h_t + \gamma_k) - I_{a_k}(h_t + \beta_k, t - h_t + \gamma_k)}{Z_k}}_{\text{Truncation Correction}}\enspace.
\end{equation}
Note that this formalism requires Equation~\ref{eqn:criterion_loss} to be updated to the form seen within Algorithm~\ref{alg:training}. Based upon this, the total wealth $W_t(p_0)$ is then the weighted sum of the localized likelihood ratios %
\begin{equation}
    E_k(p_0) = \frac{m_k(h_t, t; \mathcal{R}_k)}{p_0^{h_t} (1 - p_0)^{t-h_t}} \quad \text{and} \quad W_{t}^{meta}(p_0) = \sum_{k=1}^{K} w_k E_k(p_0)\enspace.
\end{equation}

\paragraph{Rejection Martingales and Bankruptcy Exits} The truncated beta framework enables both heuristic and mathematically rigorous exit strategies. Under the Hybrid Support mode, we concentrate a subset of mixture heads on the robust interval $[0.5, 1.0]$. If the true success probability $p$ is significantly below $0.5$, the likelihood ratios for these robust components will exponentially decay.%

At every check interval $B$, we monitor if the total wealth $W_t(p_0)$ at $p_0 = 0.5$ falls below a failure threshold $\epsilon_{fail}=0.1$---a bankruptcy exit. If this occurs, the certifier immediately halts and rejects the sample. While this may result in rare false negatives for samples very near the decision boundary, it functions as a high-speed rejection filter that preserves global compute. This design is predicated upon a core principle of certified robustness: that conservative rejections are vastly preferable to certifications that are either computationally expensive or statistically invalid. In practice we require the process to run for $400$ samples before triggering an early-rejection.

\paragraph{The Safety Anchor: Guaranteeing Survival} To guard against bankruptcy caused by misspecification of the meta-prior, we supplement the meta-learned mixture with a global safety anchor, allocating a small fixed weight $w_{safety}=0.01$ to a KT prior. This ensures the E-process remains robust even if the meta-learned components decay, by defining that
\begin{equation}\label{eqn:anchor_mixture}
  W_t(p_0) = w_{safety} E_{safety}(p_0) + (1 - w_{safety}) W_t^{meta}(p_0)\enspace,
\end{equation}
where $E_{safety}(p_0)$ is the E-value generated by a $\text{Beta}(0.5, 0.5)$ prior over the full support $[0, 1]$. Here $E_{safety}(p_0) = m_{KT}(h_t, t) / [p_0^{h_t} (1 - p_0)^{t-h_t}]$ and $m_{KT}$ is the integrated likelihood under the KT prior. This hybrid strategy provides a statistical safety net, where convergence can be reached even in cases where the meta-learners predictions would otherwise compromise convergence.%

\section{The Certification Engine: Termination Policies}\label{sec:termination}

While the structural design of the E-process outlined in the preceding subsection ensures safety, efficiency is realized through active termination policies. At each check interval $B$, we solve for the LCB $\underline{p_t}$ using the Brent-Dekker method~\citep{brent1971algorithm} to provide a continuous radius estimate $r_t = \sigma \Phi^{-1}(\underline{p_t})$. Based upon this, we consider two early-stopping frameworks.

\paragraph{Precision-Based Stopping} We implement a heuristic that defines a time-varying precision threshold $\epsilon_t$. We terminate the sequence if the gap between the Maximum Likelihood radius ($r_{mle}$) and the certified radius ($r_{lcb}$) satisfies $(r_{mle} - r_{lcb}) \le \epsilon_t$, yielding a base threshold%
\begin{equation}\label{eqn:stopping}
  \epsilon_t^{base} = \Delta \cdot \left[ \epsilon_{start} - (\epsilon_{start} - \epsilon_{end}) \cdot \frac{t}{N_{max} - N_{sel}} \right]\enspace,
\end{equation}
where $\Delta$ is an aggression factor (typically $1.2$), $\epsilon_{start}=0.1$, and $\epsilon_{end}=0.042$. This allows for rapid early exits on unambiguous samples while ensuring tight bounds for marginal cases. To support application-specific requirements, we can further modulate this threshold with a specialization bias $b(r_{mle})$---for instance, requiring higher precision ($b < 1$) in a target radius zone while allowing relaxed exits ($b > 1$) elsewhere (see Section~\ref{sec:results} for further details).

\paragraph{Adversarial and Plateau Exits} We prioritize rapid rejection for non-robust samples ($p < 0.5$). If $W_t(0.5) \geq 1/\alpha$ and the empirical mean $\hat{p}_{mle} < 0.5$, it implies the Upper Confidence Bound is less than $0.5$, and the sample is rejected. Furthermore, we monitor the Radius Velocity: if $r_{lcb}$ does not improve by more than $5\%$ over four consecutive intervals, the system triggers a certification based upon the current LCB, effectively rejecting the sample if the accumulated evidence remains below the robustness threshold, and eschewing further calculations to preserve the computational budget.

\paragraph{Algorithm} The complete operational flow of our sequential radius estimation is synthesized in Algorithms~~\ref{alg:training} and \ref{alg:certification} in Appendix~\ref{app:algorithms}. This procedure integrates bespoke prior initialization with anytime-valid monitoring, allowing for a dynamic trade-off between certification tightness and computational budget that automatically adapts to the difficulty of the input.

\section{Results and Discussion}\label{sec:results}

While raw efficiency gains are compelling, it is important to address a fundamental question: why favor an anytime-valid approach over a fixed-horizon Clopper-Pearson bound, which is inherently tighter for a given $N$? After all, there is no free lunch when it comes to anytime validity. However, as we will now argue, the advantages are two-fold: the \textbf{ability to dynamically triage samples} %
and the capacity to construct \textbf{application-aware termination conditions}.

\begin{figure}[t]
    \centering
    \includegraphics[width=\textwidth]{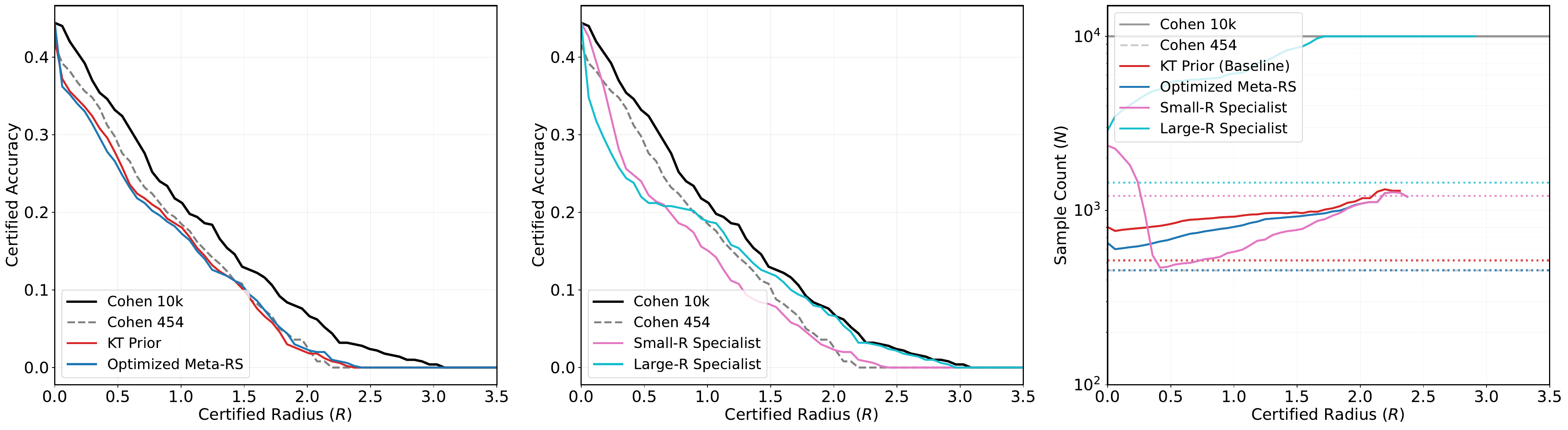}
    \caption{\textbf{Specialist Triage via Radius-Biased Stopping for CIFAR-10 at $\sigma=1.0$.} (Left) Generalist accuracy parity with Cohen-10k using 22$\times$ fewer samples. (Middle) Specialist accuracy recovery in targeted zones ($R < 0.5$ or $R > 1.0$). (Right) Corresponding \textbf{Sample Count ($N$)} across radii, showing the significant compute reduction outside target zones. Metrics metrics represent the average for samples with a radius larger than $r$, following Equation~\ref{eqn:certified_accuracy}, except the dashed lines in (Right), which represent the average counts \emph{including rejected samples}.}
    \label{fig:exemplar}
\end{figure}

We first evaluate the performance of our framework using the \texttt{Meta-1-hybrid} on a classic certification analysis, as shown in Figure~\ref{fig:exemplar} (Left). This approach provides similar certification dynamics to the $N=10,000$ Cohen baseline (hereafter `Cohen-10k') while requiring an order of magnitude fewer samples. As the average number of samples required by Meta-1-hybrid was $N=454$, we also test Cohen against the same number of samples. In this case the anytime-valid bounds induce the slight (and expected) decrease in certification performance relative to the tighter bounds of \citet{cohen2019certified}. However, we stress that \emph{a priori} knowledge of this number of samples is impossible. When exploring the comprehensive set of results in Table~\ref{tab:full_results}, the Meta-RS approach typically reduces the average samples required by $8$ to $15\%$, relative to KT, with consistent gains in the high-noise regimes, while also producing an up to $4 \%$ tighter bounds for ImageNet.

The true strength of anytime-validity is revealed when stopping conditions are tuned to specific downstream regions. In many applications, the importance of an input's radius may depend upon its scale. Such dynamics are likely in safety critical deployments, where it may be desirable to devote computational resources to refining small radius certifications (to maximize the accuracy of these certifications),  or potentially on larger-radius regions, if the small-radius samples are to be subjected to manual verification irrespective of their final radii.  

To model the implications of such dynamics, we implement \textbf{Small-R} (focusing compute upon $R < 0.5$) and \textbf{Large-R Specialists} (focusing compute upon $R > 1.0$)---details of how these are implemented can be found in Appendix~\ref{app:radius-specialized}. As seen in Figure~\ref{fig:exemplar} (Middle), by biasing the stopping epsilon $\epsilon_t$ toward targeted regions, we can perfectly recover the Cohen-10k accuracy curve in the region of interest while sacrificing precision elsewhere to save compute. These specialists effectively reduce the sample budget to a minimal rejection floor of \textbf{$\sim$300 samples} for inputs outside their target zones. This represents a \textbf{33$\times$ speedup} over the baseline for non-robust %
samples, focusing computational attention exclusively on the samples that matter.%

\subsection{Early Rejection}

An alternative perspective on early-stopping is that it should only be applied to \textbf{reject non-robust samples}, with the total sample count otherwise being let to run to the full $10,000$ sample horizon. Under these conditions, we only allow the model to exit if its UCB falls below $0.5$ or its wealth process falls below $0.1$ (bankruptcy). Such an approach is impossible under traditional certification workflows, which must still expend their full computational budget on non-robust samples. 

Table~\ref{tab:early_exit_efficiency} demonstrates that orders of magnitude decreases in computational cost for rejections only elicits a small decrease in certification performance, while requiring an up to $45\times$ speedup for sample rejections. That this occurs with only a minor penalty in terms of the achievable radius---stemming from the anytime-valid tax---demonstrates the utility of this approach for allocating computational load to the samples that matter.

\begin{table}[h]
\centering
\small
\setlength{\tabcolsep}{4pt}
\caption{\textbf{Performance of the Efficiency Champion Model} (Meta-1-Dynamic-Margin - see Appendix~\ref{app:champion_selection}) vs KT Prior baseline, when samples are allowed to run to $10,000$ if robust. Cert (\%): Certified accuracy at the full horizon; $\delta r$: Mean \% discrepancy in certified radius relative to the Cohen-10k ground truth; Exit: System-wide average sample latency (including early rejections).}
\label{tab:early_exit_efficiency}
\begin{tabular}{llccccc}
\toprule
Dataset & Sigma & Cert (\%) & $\delta r_{Meta}$ (\%) & $\delta r_{KT}$ (\%) & Exit Meta & Exit KT \\
\midrule
\multirow{3}{*}{MNIST} & 0.25 & 99.20 & 1.89 & 5.08 & 200 & 350.00 \\
 & 0.50 & 98.80 & 2.20 & 4.46 & 433.33 & 283.33 \\
 & 1.00 & 91.20 & 2.97 & 4.01 & 1659.09 & 402.27 \\
 \midrule
\multirow{3}{*}{CIFAR-10} & 0.25 & 78.80 & 3.22 & 4.87 & 1003.77 & 322.64 \\
 & 0.50 & 62.60 & 4.31 & 5.60 & 862.57 & 326.74 \\
 & 1.00 & 44.40 & 4.21 & 5.56 & 788.129 & 308.63 \\
 \midrule
\multirow{3}{*}{ImageNet} & 0.25 & 76.20 & 2.47 & 5.18 & 232.773 & 232.77 \\
 & 0.50 & 73.80 & 2.49 & 4.90 & 263.36 & 263.36 \\
 & 1.00 & 72.40 & 3.33 & 5.28 & 534.783 & 288.41 \\
\bottomrule
\end{tabular}
\end{table}

\subsection{Case-Study: Operational Triage Framework}\label{sec:triage}

To further consider the potential for E-values to support dynamic triage, we consider a scenario where the exact values of the certification hold little value, relative to the positioning of the certification within a region defined by its perceived risk. Under fixed-horizon methods, adapting to such a downstream scenario would be impossible---the sample budget must be committed to all inputs. In contrast, E-value certifications halt as soon as the confidence interval $[LCB_t, UCB_t]$ is entirely contained within one of these buckets. To test this, we consider the partitioning: 
\begin{itemize}[leftmargin=*, noitemsep, topsep=0pt]
    \item \textbf{Bucket A (Non-Robust):} $r = 0.0$. Immediate rejection or human intervention.
    \item \textbf{Bucket B (Low Robustness):} $0.0 < r \le 1.0$. Automated scrutiny.
    \item \textbf{Bucket C (Medium Robustness):} $1.0 < r \le 1.5$. Monitored deployment.
    \item \textbf{Bucket D (High Robustness):} $r > 1.5$. Autonomous deployment.
\end{itemize}

To explore the performance under this operational triage framework, we tested \citet{cohen2019certified} at $10,000$ samples, a KT Prior Method-of-Mixtures approach with a constant $\alpha=0.01$, and two application specific approaches. The Tiered Spec Race assigns each bin a specific budget $\{\alpha_A, \dots, \alpha_D\} = \{0.05, 0.05, 0.025, 0.01\}$, representing the levels of tolerance of a false-positive in each category, and performs four independent E-processes for each operational bucket. Samples are assigned to bucket $i$ at timestep $T$ if its LCB and UCB are contained within the associated bin with $M_i(X_{1:T}) \ge 1/\alpha'_i$, where $\alpha'_i = \alpha_i \cdot (0.2 + 0.8 \cdot w_{meta, i})$.%

The Relaxed Tiered Cascade follows the above process, however, if samples remain unresolved after $400$ samples the system attempts to prove proximity to any bin boundary $p_b \in \{0.5, 0.8413, 0.9332\}$ (corresponding to $r = \{0, 1, 1.5\}$), through a dual-rejection process considering  $H_0^-: p = p_b - \epsilon$ and $H_0^+: p = p_b + \epsilon$, using an $\epsilon=0.05$. If $M(X; p_b - \epsilon) \ge 1/\alpha$ and $M(X; p_b + \epsilon) \ge 1/\alpha$, the sample is mathematically trapped within the $[p_b - \epsilon, p_b + \epsilon]$ buffer. If this is achieved, the sample is classified to bin, but provided a secondary label denoting boundary ambiguity. This approach prevents allocating redundant samples to distinguishing infinitesimally close probabilities, such a $p=0.841$ from $p=0.842$. Table~\ref{tab:triage} demonstrates that the Meta-learner's ability to partition risk and predict sample-specific density leads to an up to %
\textbf{2.4$\times$ speedup} over the KT Prior baseline. %

\begin{table}[h]
\centering
\small
\setlength{\tabcolsep}{4pt}
\caption{\textbf{Operational Triage Performance} (CIFAR-10, $\sigma=1.0$, $N=3000$). Success is defined as proving containment within the correct triage bucket.}
\label{tab:triage}
\begin{tabular}{lrrcc}
\toprule
Strategy & Avg Samples & Avg Bin A & Success (\%) & Definitive (\%) \\
\midrule
Cohen-10k (Ref) & 10,000.0 & 10,000.0 & 100.0\% & 100.0\% \\
KT Prior (Fixed) & 1,733.2 & 948.8 & 100.0\% & 90.8\% \\
Tiered Spec Race & 1,448.8 & 988.4 & 99.9\% & 92.5\% \\
Relaxed Tiered Cascade & \textbf{710.5} & \textbf{480.0} & 99.9\% & 92.5\% \\
\bottomrule
\end{tabular}
\end{table}

\subsection{Ablation}
\paragraph{Component-Wise Efficiency Attribution} To understand the drivers of the 20x--45x speedups observed in our champions, we perform an attribution analysis, the results of which are further detailed in Figure~\ref{fig:ablation_5}. While Precision-Based Stopping provides the bulk of the speedup for certifiable samples, we find that the combination of Bankruptcy and UCB exits is critical for minimizing the costs associated with rejecting non-robust samples, enabling best in class performance. %

\paragraph{Statistical Universality and Zero-Shot Transfer}
A key concern for meta-learned certifiers is dataset dependency. To evaluate generalization, we employ ImageNet to test a Meta-Learner trained exclusively on CIFAR-10. Figure~\ref{fig:ablation_4} demonstrates that there is some correlation in termination latencies. This suggests the meta-learner captures universal properties of the classifiers rather than dataset-specific artifacts, enabling zero-shot deployment on new architectures.

\paragraph{The Precision-Efficiency Pareto Frontier}
Anytime-validity transforms certified robustness from a binary outcome into a tunable resource. As detailed in Figure~\ref{fig:ablation_7}, sample complexity grows only logarithmically with $\epsilon$ over a limited window. This suggests that there is some ability to reliably trade compute for tighter bounds, or conversely, to relax precision for real-time throughput.

\section{Conclusion}

In this work, we demonstrated that randomized smoothing can be transformed from a static, computationally exhaustive process into a dynamic, application-aware certification pipeline optimized for \textbf{strategic resource management}. While computational latency has always presented significant concerns for certified robustness, our E-value meta-learner produces substantial empirical gains in this space, without compromising robustness. Achieving a $89\%$ reduction in computational load relative to standard certification workflows, or $15\%$ relative to the KT prior, validates the utility of our meta-learning approach. 

However, headline computational metrics do not tell the full story of this work. Our framework demonstrates that it is possible to dynamically balance robustness with compute on a sample-by-sample basis, through early rejections or downstream-focused triage. While anytime-validity incurs a marginal cost regarding the realizable tightness of bounds, our results highlight the overwhelming utility of transitioning to meta-learned E-values. The true value of this approach lies not just in its computational efficiency, but in proving that dynamic, application-specific triage is finally possible, marking a definitive leap toward the real-world deployment of certified machine learning.

\newpage

\section*{Acknowledgments}

This work was supported by the Australian Defence Science and Technology (DST) Group via the Advanced Strategic Capabilities Accelerator (ASCA) program.

\section*{Impact Statement}

This work explores the potential for improving the computational efficiency of mechanisms for achieving Certified Robustness, with the aim of significantly enhancing the range of applicability for these systems. Improving the robustness of models, and lowering the energy consumption of our validation processes, has clear societal implications, especially as we move to a landscape where AI is increasingly integrated into society. However, with that said, we feel that there are two key societal concerns with pursuing robustness research, which we must note here. 

The first of which is that there are some applications where a lack of robustness in a model may be positive. In a world where broad scale surveillance is increasingly normalized, it may well be the case that adversarial attacks may induce privacy, creating a net public good. 

The second relates to how works like this position risk and harm. A common precept within the Adversarial Machine Learning community is to assume a particular threat model, with the nature of academic comparisons often incentivizing us to then follow in the footsteps of those who came before us. However, in doing so, we inadvertently create---and, crucially, present---a myopic view of the risk landscape. In essence, we portray to practitioners that risk is concentrated within the areas in which we act as a community, when our investigations may be more motivated by historic alignment to academic norms and mathematical convenience. This work considers $\ell_2$ perturbations, which while aligned with acoustic threat models, still represent a restriction relative to the overall threat landscape. 

We emphasize the above point not just for the risks of erroneous portrayals of risk to practitioners, but also because our focus on these spaces inherently biases real attacker behavior away from these threat models. After all, if an attacker understands that an $\ell_2$ threat model is likely defended against, they're naturally incentivized to consider an alternative pathway for model manipulation. 

With these points made, we still believe that research into defenses, and in particular certified defenses, induce a net societal gain. Improving robustness to natural or adversarial perturbations will improve the performance of systems that are already one of the dominant access portals for AI within the community.

\bibliographystyle{icml2026}
\bibliography{references}

@article{szegedy2013intriguing,
    author = {Szegedy, Christian and Zaremba, Wojciech and Sutskever, Ilya and Bruna, Joan and Erhan, Dumitru and Goodfellow, Ian and Fergus, Rob},
    journal = {arXiv preprint arXiv:1312.6199},
    title = {{I}ntriguing {P}roperties of {N}eural {N}etworks},
    year = {2013}
}

@article{goodfellow2014explaining,
    author = {Goodfellow, Ian J and Shlens, Jonathon and Szegedy, Christian},
    journal = {arXiv preprint arXiv:1412.6572},
    title = {{E}xplaining and {H}arnessing {A}dversarial {E}xamples},
    year = {2014}
}

@inproceedings{madry2017towards,
    author = {Aleksander Madry and Aleksandar Makelov and Ludwig Schmidt and Dimitris Tsipras and Adrian Vladu},
    booktitle = {International Conference on Learning Representations},
    title = {{T}owards {D}eep {L}earning {M}odels {R}esistant to {A}dversarial {A}ttacks},
    year = {2018}
}

@inproceedings{cullen25a,
    author = {Cullen, Andrew Craig and Montague, Paul and Erfani, Sarah Monazam and Rubinstein, Benjamin I. P.},
    booktitle = {Proceedings of the 42nd International Conference on Machine Learning},
    month = {13--19 Jul},
    pages = {81185--81198},
    publisher = {PMLR},
    series = {Proceedings of Machine Learning Research},
    title = {{P}osition: {C}ertified {R}obustness {D}oes {N}ot ({{Y}}et) {I}mply {M}odel {S}ecurity},
    volume = {267},
    year = {2025}
}

@inproceedings{cohen2019certified,
    author = {Cohen, Jeremy and Rosenfeld, Elan and Kolter, Zico},
    booktitle = {International Conference on Machine Learning},
    organization = {PMLR},
    pages = {1310--1320},
    title = {{C}ertified {{A}}dversarial {{R}}obustness via {{R}}andomized {{S}}moothing},
    year = {2019}
}

@inproceedings{lecuyer2019certified,
  title={Certified {R}obustness to {A}dversarial {E}xamples with {D}ifferential {P}rivacy},
  author={Lecuyer, Mathias and Atlidakis, Vaggelis and Geambasu, Roxana and Hsu, Daniel and Jana, Suman},
  booktitle={2019 {IEEE} {S}ymposium on {S}ecurity and {P}rivacy ({SP})},
  pages={656--672},
  year={2019},
  organization={{IEEE} {C}omputer {S}ociety}
}

@article{cullen2022double,
  title={Double {B}ubble, {T}oil and {T}rouble: {E}nhancing {C}ertified {R}obustness through {T}ransitivity},
  author={Cullen, Andrew C. and Montague, Paul and Liu, Shijie and Erfani, Sarah M. and Rubinstein, Benjamin I.P.},
  journal={Advances in {N}eural {I}nformation {P}rocessing {S}ystems},
  volume={35},
  pages={19099--19112},
  year={2022}
}

@article{voravcek2024treatment,
  title={Treatment of {S}tatistical {E}stimation {P}roblems in {R}andomized {S}moothing for {A}dversarial {R}obustness},
  author={Vor{\'a}{\v{c}}ek, V{\'a}clav},
  journal={Advances in {N}eural {I}nformation {P}rocessing {S}ystems},
  volume={37},
  pages={133464--133486},
  year={2024}
}

@article{clopper1934use,
  title={The {U}se of {C}onfidence or {F}iducial {L}imits {I}llustrated in the case of the {B}inomial},
  author={Clopper, Charles J and Pearson, Egon S},
  journal={Biometrika},
  volume={26},
  number={4},
  pages={404--413},
  year={1934},
  publisher={JSTOR}
}

@article{doob1940regularity,
  title={Regularity {P}roperties of {C}ertain {F}amilies of {C}hance {V}ariables},
  author={Doob, Joseph L},
  journal={Transactions of the American Mathematical Society},
  volume={47},
  number={3},
  pages={455--486},
  year={1940}
}

@book{ville1939etude,
  title={Etude {C}ritique de la {N}otion de {C}ollectif},
  author={Ville, Jean},
  volume={3},
  year={1939},
  publisher={Gauthier-Villars Paris}
}

@article{brent1971algorithm,
  title={An {A}lgorithm with {G}uaranteed {C}onvergence for {F}inding a {Z}ero of a {F}unction},
  author={Brent, Richard P.},
  journal={The {C}omputer {J}ournal},
  volume={14},
  number={4},
  pages={422--425},
  year={1971},
  publisher={Oxford University Press}
}

@article{krichevsky1981performance,
  title={The {P}erformance of {U}niversal {E}ncoding},
  author={Krichevsky, Raphail and Trofimov, Victor},
  journal={{IEEE} {T}ransactions on {I}nformation {T}heory},
  volume={27},
  number={2},
  pages={199--207},
  year={1981},
  publisher={{IEEE}}
}

@article{xie2000asymptotic,
  title={Asymptotic {M}inimax {R}egret for {D}ata {C}ompression, {G}ambling, and {P}rediction},
  author={Xie, Qun and Barron, Andrew R},
  journal={IEEE Transactions on Information Theory},
  volume={46},
  number={2},
  pages={431--445},
  year={2000}
}

@book{shafer2019game,
  title={Game-{T}heoretic {F}oundations for {P}robability and {F}inance},
  author={Shafer, Glenn and Vovk, Vladimir},
  year={2019},
  publisher={John Wiley \& Sons}
}

@article{ramdas2023game,
  title={Game-{T}heoretic {S}tatistics and {S}afe {A}nytime-{V}alid {I}nference},
  author={Ramdas, Aaditya and Gr{\"u}nwald, Peter and Vovk, Vladimir and Shafer, Glenn},
  journal={Statistical Science},
  volume={38},
  number={4},
  pages={576--601},
  year={2023},
  publisher={Institute of Mathematical Statistics}
}

@article{waudby2024estimating,
  title={Estimating {M}eans of {B}ounded {R}andom {V}ariables by {B}etting},
  author={Waudby-Smith, Ian and Ramdas, Aaditya},
  journal={Journal of the {R}oyal {S}tatistical {S}ociety: {S}eries B-{S}tatistical {M}ethodology},
  volume={86},
  number={1},
  year={2024},
  pages={1-27},
  publisher={Oxford {U}niversity {P}ress}
}

@inproceedings{grunwald2020safe,
  title={Safe {T}esting},
  author={Gr{\"u}nwald, Peter and de Heide, Rianne and Koolen, Wouter M},
  booktitle={2020 Information {T}heory and {A}pplications workshop (ITA)},
  pages={1--54},
  year={2020},
  organization={{IEEE}}
}

@article{vovk2021values,
  title={E-values: {C}alibration, {C}ombination and {A}pplications},
  author={Vovk, Vladimir and Wang, Ruodu},
  journal={The Annals of Statistics},
  volume={49},
  number={3},
  pages={1736--1754},
  year={2021},
  publisher={Institute of Mathematical Statistics}
}

@inproceedings{salman2019convex,
  title={A {C}onvex {R}elaxation {B}arrier to {T}ight {R}obustness {V}erification of {N}eural {N}etworks},
  author={Salman, Hadi and Yang, Greg and Zhang, Huan and Hsieh, Cho-Jui and Zhang, Pengchuan},
  booktitle={Advances in Neural Information Processing Systems},
  year={2019}
}

@inproceedings{mirman2018differentiable,
  title={Differentiable {A}bstract {I}nterpretation for {P}rovably {R}obust {N}eural {N}etworks},
  author={Mirman, Matthew and Gehr, Timon and Vechev, Martin},
  booktitle={International Conference on Machine Learning},
  pages={3578--3586},
  year={2018},
  organization={PMLR}
}

@inproceedings{weng2018towards,
  title={Towards {F}ast {C}omputation of {C}ertified {R}obustness for {R}e{LU} {N}etworks},
  author={Weng, Lily and Zhang, Huan and Chen, Hongge and Song, Zhao and Hsieh, Cho-Jui and Daniel, Luca and Boning, Duane and Dhillon, Inderjit},
  booktitle={International Conference on Machine Learning},
  pages={5276--5285},
  year={2018},
  organization={PMLR}
}

@inproceedings{zhang2018efficient,
  title={Efficient {N}eural {N}etwork {R}obustness {C}ertification with {G}eneral {A}ctivation {F}unctions},
  author={Zhang, Huan and Weng, Tsui-Wei and Chen, Pin-Yu and Hsieh, Cho-Jui and Daniel, Luca},
  booktitle={Neural Information Processing Systems (NeurIPS)},
  year={2018}
}

@article{singh2019abstract,
  title={An {A}bstract {D}omain for {C}ertifying {N}eural {N}etworks},
  author={Singh, Gagandeep and Gehr, Timon and P{\"u}schel, Markus and Vechev, Martin},
  journal={Proceedings of the ACM on Programming Languages},
  volume={3},
  number={POPL},
  pages={1--30},
  year={2019},
  publisher={ACM New York, NY, USA}
}

@inproceedings{mohapatra2020towards,
  title={Towards {V}erifying {R}obustness of {N}eural {N}etworks against a family of {S}emantic {P}erturbations},
  author={Mohapatra, Jeet and Weng, Tsui-Wei and Chen, Pin-Yu and Liu, Sijia and Daniel, Luca},
  booktitle={Proceedings of the IEEE/CVF Conference on Computer Vision and Pattern Recognition},
  pages={244--252},
  year={2020}
}

@inproceedings{lyu2021towards,
  title={Towards {E}valuating and {T}raining {V}erifiably {R}obust {N}eural {N}etworks},
  author={Lyu, Zhaoyang and Guo, Minghao and Wu, Tong and Xu, Guodong and Zhang, Kehuan and Lin, Dahua},
  booktitle={Proceedings of the IEEE/CVF Conference on Computer Vision and Pattern Recognition},
  pages={4308--4317},
  year={2021}
}

@article{xu2020automatic,
  title={Automatic {P}erturbation {A}nalysis for {S}calable {C}ertified {R}obustness and {B}eyond},
  author={Xu, Kaidi and Shi, Zhouxing and Zhang, Huan and Wang, Yihan and Chang, Kai-Wei and Huang, Minlie and Kailkhura, Bhavya and Lin, Xue and Hsieh, Cho-Jui},
  journal={Advances in Neural Information Processing Systems},
  volume={33},
  year={2020}
}

@article{wang2021beta,
  title={Beta-{CROWN}: {E}fficient {B}ound {P}ropagation with {P}er-{N}euron {S}plit {C}onstraints for {N}eural {N}etwork {R}obustness {V}erification},
  author={Wang, Shiqi and Zhang, Huan and Xu, Kaidi and Lin, Xue and Jana, Suman and Hsieh, Cho-Jui and Kolter, J Zico},
  journal={Advances in Neural Information Processing Systems},
  volume={34},
  year={2021}
}

@article{chiang2020certified,
  title={Certified {D}efenses for {A}dversarial {P}atches},
  author={Chiang, Ping-yeh and Ni, Renkun and Abdelkader, Ahmed and Zhu, Chen and Studer, Christoph and Goldstein, Tom},
  journal={arXiv preprint arXiv:2003.06693},
  year={2020}
}

@article{levine2020randomized,
  title={(De){R}andomized {S}moothing for {C}ertifiable {D}efense against {P}atch {A}ttacks},
  author={Levine, Alexander and Feizi, Soheil},
  journal={Advances in Neural Information Processing Systems},
  volume={33},
  pages={6465--6475},
  year={2020}
}

@inproceedings{horvath2022boosting,
  title={Boosting Randomized Smoothing with Variance Reduced Classifiers},
  author={Horv{\'a}th, Mikl{\'o}s Z and Mueller, Mark Niklas and Fischer, Marc and Vechev, Martin},
  booktitle={International Conference on Learning Representations},
  year={2022}
}

@inproceedings{chen2022input,
  title={Input-{S}pecific {R}obustness {C}ertification for {R}andomized {S}moothing},
  author={Chen, Ruoxin and Li, Jie and Yan, Junchi and Li, Ping and Sheng, Bin},
  booktitle={Proceedings of the AAAI Conference on Artificial Intelligence},
  volume={36},
  number={6},
  pages={6295--6303},
  year={2022}
}

@article{howard2021time,
  title={Time-{U}niform, {N}onparametric, {N}onasymptotic {C}onfidence {S}equences},
  author={Howard, Steven R and Ramdas, Aaditya and McAuliffe, Jon and Sekhon, Jasjeet},
  journal={The Annals of Statistics},
  volume={49},
  number={2},
  pages={1055--1080},
  year={2021},
  publisher={JSTOR}
}

@inproceedings{li2018certified,
  author    = {Bai Li and
               Changyou Chen and
               Wenlin Wang and
               Lawrence Carin},
  title     = {Certified {A}dversarial {R}obustness with {A}dditive {N}oise},
  booktitle = {Advances in Neural Information Processing Systems},
  pages     = {9459--9469},
  volume={32},
  year      = {2019},
  organization={NeurIPS}
}

@inproceedings{dwork2006calibrating,
  title={Calibrating {N}oise to {S}ensitivity in {P}rivate {D}ata {A}nalysis},
  author={Dwork, Cynthia and McSherry, Frank and Nissim, Kobbi and Smith, Adam},
  booktitle={Theory of Cryptography Conference},
  pages={265--284},
  year={2006},
  organization={Springer},
  series={TCC}
}

@inproceedings{shi2023formal,
  title={Formal {V}erification for {N}eural {N}etworks with {G}eneral {N}onlinearities via {B}ranch-{A}nd-{B}ound},
  author={Shi, Zhouxing and Jin, Qirui and Zhang, Huan and Kolter, Zico and Jana, Suman and Hsieh, Cho-Jui},
  booktitle={2nd Workshop on Formal Verification of Machine Learning (WFVML 2023)},
  year={2023}
}

@inproceedings{hein2017formal,
  title = {Formal {G}uarantees on the {R}obustness of a {C}lassifier {A}gainst {A}dversarial {M}anipulation},
  author = {Hein, Matthias and Andriushchenko, Maksym},
booktitle = {Advances in Neural Information Processing Systems},
  series = {NeurIPS},
  year = {2017},
  volume={30}
}

@inproceedings{tsuzuku2018lipschitz,
  title={Lipschitz-{M}argin {T}raining: {S}calable {C}ertification of {P}erturbation {I}nvariance for {D}eep {N}eural {N}etworks},
  author={Tsuzuku, Yusuke and Sato, Issei and Sugiyama, Masashi},
  booktitle={Advances in Neural Information Processing Systems},
  year={2018},
  volume={31},
  organization={NeurIPS}
}

@inproceedings{leino2021globally,
  title={Globally-{R}obust {N}eural {N}etworks},
  author={Leino, Klas and Wang, Zifan and Fredrikson, Matt},
  booktitle={International Conference on Machine Learning},
  pages={6212--6222},
  year={2021},
  organization={PMLR}
}

@inproceedings{salman2019provably,
 author = {Salman, Hadi and Li, Jerry and Razenshteyn, Ilya and Zhang, Pengchuan and Zhang, Huan and Bubeck, Sebastien and Yang, Greg},
 booktitle = {Advances in Neural Information Processing Systems},
 title= {Provably {R}obust {D}eep {L}earning via {A}dversarially {T}rained {S}moothed {C}lassifiers},
 year={2019},
 volume={32},
 pages={11292--11303},
 organization={NeurIPS}
}

@inproceedings{johari2017peeking,
  title={Peeking at {A}/{B} {T}ests: {W}hy it matters, and what to do about it},
  author={Johari, Ramesh and Koomen, Pete and Pekelis, Leonid and Walsh, David},
  booktitle={Proceedings of the 23rd ACM SIGKDD International Conference on Knowledge Discovery and Data Mining},
  pages={1517--1525},
  year={2017}
}

@inproceedings{zhai2020macer,
  title={Macer: {A}ttack-{F}ree and {S}calable {R}obust {T}raining via {M}aximizing {C}ertified {R}adius},
  author={Zhai, Runtian and Dan, Chen and He, Di and Zhang, Huan and Gong, Boqing and Ravikumar, Pradeep and Hsieh, Cho-Jui and Wang, Liwei},
  booktitle={International Conference on Learning Representations},
  year={2020},
}

@article{shafer2019language,
  title={The {L}anguage of {B}etting as a {S}trategy for {S}tatistical and {S}cientific {C}ommunication},
  author={Shafer, Glenn},
  journal={arXiv preprint arXiv:1903.06991},
  year={2019}
}

@article{lai1976confidence,
  title={On {C}onfidence {S}equences},
  author={Lai, Tze Leung},
  journal={The Annals of Statistics},
  pages={265--280},
  year={1976},
  publisher={JSTOR}
}

@article{kelly1956new,
  title={A {N}ew {I}nterpretation of {I}nformation {R}ate},
  author={Kelly, John L},
  journal={{T}he {B}ell {S}ystem {T}echnical {J}ournal},
  volume={35},
  number={4},
  pages={917--926},
  year={1956},
  publisher={Nokia Bell Labs}
}

@inproceedings{gershman2014amortized,
  title={Amortized {I}nference in {P}robabilistic {R}easoning},
  author={Gershman, Samuel and Goodman, Noah},
  booktitle={Proceedings of the {A}nnual {M}eeting of the {C}ognitive {S}cience {S}ociety},
  volume={36},
  number={36},
  year={2014}
}

@article{lecun1998gradient,
  title={Gradient-{B}ased {L}earning {A}pplied to {D}ocument {R}ecognition},
  author={LeCun, Yann and Bottou, L{\'e}on and Bengio, Yoshua and Haffner, Patrick},
  journal={Proceedings of the IEEE},
  volume={86},
  number={11},
  pages={2278--2324},
  year={1998},
  publisher={Ieee}
}

@techreport{krizhevsky2009learning,
title={Learning {M}ultiple {L}ayers of {F}eatures from {T}iny {I}mages},
  author={Krizhevsky, Alex and Hinton, Geoffrey and others},
  year={2009},
  institution={University of Toronto}
}

@inproceedings{NEURIPS2019_9015,
title = {PyTorch: {A}n {I}mperative {S}tyle, {H}igh-{P}erformance {D}eep {L}earning {L}ibrary},
author = {Paszke, Adam and Gross, Sam and Massa, Francisco and Lerer, Adam and Bradbury, James and Chanan, Gregory and Killeen, Trevor and Lin, Zeming and Gimelshein, Natalia and Antiga, Luca and Desmaison, Alban and Kopf, Andreas and Yang, Edward and DeVito, Zachary and Raison, Martin and Tejani, Alykhan and Chilamkurthy, Sasank and Steiner, Benoit and Fang, Lu and Bai, Junjie and Chintala, Soumith},
booktitle = {Advances in Neural Information Processing Systems},
editor = {H. Wallach and H. Larochelle and A. Beygelzimer and F. d'Alch\'{e}-Buc and E. Fox and R. Garnett},
pages = {8024--8035},
year = {2019},
volume={32},
organization={NeurIPS}
}

@article{russakovsky2015imagenet,
  title={Image{N}et {L}arge {S}cale {V}isual {R}ecognition {C}hallenge},
  author={Russakovsky, Olga and Deng, Jia and Su, Hao and Krause, Jonathan and Satheesh, Sanjeev and Ma, Sean and Huang, Zhiheng and Karpathy, Andrej and Khosla, Aditya and Bernstein, Michael and others},
  journal={International Journal of Computer Vision},
  volume={115},
  number={3},
  pages={211--252},
  year={2015},
  publisher={Springer}
}

\appendix

\section{Expanded Related Work}\label{sec:extended_related}

\subsection{Certification Mechanisms}

First introduced by \citet{lecuyer2019certified}, randomized smoothing based certified robustness builds upon Monte Carlo estimators of the expectation of a class prediction. While the original formulation was constructed in terms of differential privacy~\citep{dwork2006calibrating}, recent approaches have improved performance through R\'{e}nyi divergence \citep{li2018certified} and parameterising worst-case behaviors \citep{cohen2019certified, salman2019provably, cullen2022double}.

Significant research has focused on improving the underlying base classifier $f$ to be more amenable to randomized smoothing. \citet{salman2019provably} employs adversarial training to harden the model against perturbed inputs, while \textbf{MACER}~\citep{zhai2020macer} directly optimizes a robustness loss that encourages large classification margins. Our work is orthogonal to these training-time interventions; we focus on the \emph{inference-time} efficiency of the certification process itself, allowing for a 20x--45x reduction in sample complexity for \emph{any} base model, including those that have undergone the training-time interventions of MACER and \citet{salman2019provably}.

In all works, the primary metric for evaluating Randomized Smoothing is the \textbf{Certified Accuracy} at radius $r$, defined as the fraction of the test set that is both correctly classified by the smoothed model and has a certified radius $R \ge r$
\begin{equation}\label{eqn:certified_accuracy}
    \text{Acc}(r) = \mathbb{P}_{(x, y) \sim \mathcal{D}}[g(x) = y \text{ and } R(x) \ge r]
\end{equation}
where $g$ is the smoothed classifier.

\subsubsection{Interval Bound Propagation}

In the absence of probabilistic methods, conservative certificates upon the impact of norm-bounded perturbations can be constructed by way of either interval bound propagation (IBP) which propagates interval bounds through the model; or convex relaxation, which utilizes linear relaxation to construct bounding output polytopes over input bounded perturbations. In contrast to randomized smoothing, which constructs isotropic measures of $\ell_p$-robustness, interval bound propagation and its associated techniques attempt to propagate the potential influence of all possible perturbations through the model, producing an anisotropic measure of the potential response of a model to any potential perturbation~\citep{salman2019convex, mirman2018differentiable, weng2018towards, zhang2018efficient, singh2019abstract, mohapatra2020towards}. Of these, IBP is more general, while convex relaxation typically provides tighter bounds~\citep{lyu2021towards}.  

Utilizing these techniques requires introducing an augmented loss function during training to promote tight output bounds \citep{xu2020automatic}---creating significant architectural friction relative to RS style certifications, which can be applied to any model architecture. Bound propagation schemes have also, until very recently, been heavily limited in the types of network architectures that they can successfully construct bounds through, with only recent works demonstrating an applicability to a nonlinear activation functions beyond ReLU~\citep{shi2023formal}. Moreover they both exhibit a time and memory complexity that makes them infeasible for complex model architectures or high-dimensional data~\citep{wang2021beta, chiang2020certified, levine2020randomized}.

\subsubsection{Global Lipschitz}

Global Lipschitz takes an alternative approach to constructing certifications, a point that they distinguish through the framing of local and global robustness. The guarantees provided by prior works, which can take the form
\begin{equation}
    \| \mathbf{x} - \mathbf{x}' \|_p \leq \epsilon \implies F(\mathbf{x}) = F(\mathbf{x}')
\end{equation}
are considered to be local properties, that relate $\mathbf{x}$ and $\epsilon$. Lipschitz based techniques instead attempt to construct their certifications in terms of \emph{global} robustness, where
\begin{equation}
    \forall \mathbf{x}_1, \mathbf{x}_2 : \| \mathbf{x}_1 - \mathbf{x}_2 \|_p \leq \epsilon \implies F(\mathbf{x}_1) \overset{\perp}{=}  F(\mathbf{x}_2)\enspace.
\end{equation}
Here $\perp$ is the marker for an \emph{abstained} class prediction, and $c_1 \overset{\perp}{=} c_2$ denotes that either $c_1 = \perp$, $c_2 = \perp$, or $c_1 = c_2$. In essence such a form of certification involves constructing a model that has not only an infinitesimally thin decision boundary, but a margin between the regions associated with each class, where $\epsilon$ then becomes the shortest $\ell_p$ distance to span the boundary. Several attempts have been made to use Lipschitz bounds during training to promote robustness. These include constructing provable lower bounds on the norm of the input manipulation required to change classifier decisions based upon the network architecture~\citep{hein2017formal}; modifying the loss associated with logits different than the ground-truth class~\citep{tsuzuku2018lipschitz}; and GloRoNets, which add an additional logit corresponding to the predicted class at a point~\citep{leino2021globally}. While these techniques can be an order of magnitude faster than randomized smoothing, they are both less flexible---in terms of the architectures they support---and often produce smaller certifications than randomized smoothing.~\citep{leino2021globally}.

\subsection{Anytime Valid Mechanisms}\label{app:LIL}

As an alternative to betting-based martingales, anytime-valid confidence sequences can be constructed using the Law of the Iterated Logarithm (LIL). Of particular note is the \emph{stitching} construction from \citet{howard2021time}, which provides a boundary $u_t$ such that the probability of the empirical mean ever crossing $u_t$ is bounded by $\alpha$. While LIL-stitching is robust to prior mis-specification, it is generally less efficient than Mixture E-values when a reasonably accurate prior (like our Meta-Learner) is available, as it lacks the ability to aggressively allocate its convergence on specific regions of the hypothesis space. In our testing, LIL based approaches provide a small improvement on traditional \citet{cohen2019certified} style certified robustness, however, it is significantly slower to converge than our Meta-approach.

\section{Algorithms}\label{app:algorithms}

To provide further details to the processes outlined in Sections~\ref{sec:meta-learning} and \ref{sec:termination}, we present the full expanded algorithm for the training of the Meta-Learner using a Kelly-Optimal inspired process in Algorithm~\ref{alg:training}, and the overall certification approach in Algorithm~\ref{alg:certification}. 

\begin{algorithm}[t]
\caption{Meta-Learner Training (Kelly-Optimal Betting)}\label{alg:training}
\begin{algorithmic}[1]
\State \textbf{Input:} Training dataset $\mathcal{D}_{train}$, noise level $\sigma$, base classifier $f$, augmentation factor $M$, epochs $E$.
\State \textbf{Phase I: Data Collection (Offline)}
\State $\mathcal{S} \leftarrow \emptyset$
\For{each image $x^{(i)} \in \mathcal{D}_{train}$}
    \State Extract embedding $\phi^{(i)}$ and clean-image softmax $\mathbf{p}^{(i)}$.
    \State Draw $N_{max}$ samples to obtain bitstream successes $h_N^{(i)}$ and rate $p_{true}^{(i)}$.
    \State $\mathcal{S} \leftarrow \mathcal{S} \cup \{ (\phi^{(i)}, \mathbf{p}^{(i)}, h_N^{(i)}, p_{true}^{(i)}) \}$ \Comment{Store ground-truth tuples}
\EndFor
\State \textbf{Phase II: Kelly Optimization (Online)}
\State Initialize Meta-Learner $\mathcal{M}_{\theta}$.
\For{epoch $e = 1$ to $E$}
    \State Sample an index set $\mathcal{I} \subset \{1, \dots, |\mathcal{S}|\}$ of size $n$ uniformly at random.
    \For{$i \in \mathcal{I}$}
         \State $\hat{h}_{sel}^{(i)} \sim \text{Binomial}(N_{sel}, p_{true}^{(i)})$ \Comment{Sample a synthetic Phase I glimpse}
         \State $\hat{p}_{sel}^{(i)} \leftarrow \hat{h}_{sel}^{(i)} / N_{sel}$.
         \State $(\mathbf{w}^{(i)}, \mathbf{z}^{(i)}) \leftarrow \mathcal{M}_\theta(\phi^{(i)}, \mathbf{p}^{(i)}, \hat{p}_{sel}^{(i)})$. \Comment{Predict weights and raw parameters}
         \State $\boldsymbol{\beta}^{(i)}, \boldsymbol{\gamma}^{(i)} \leftarrow \text{Clamp}(\text{Softplus}(\mathbf{z}^{(i)}) + 0.5, [0.1, 500.0])$.     \Comment{Survival bias \&
     Stability}
     \EndFor
     \State \textit{// Minimize negative expected log-wealth (Kelly Loss)}
     \State $\lambda_e \leftarrow \max(1.0, 10.0 \cdot (1 - \frac{e}{0.8 E}))$ \Comment{Decay Penalty}
     \State  $\mathcal{L}(\theta) = - \frac{1}{n} \sum_{i} \left[ \log \sum_{k} w_k \cdot \frac{P(h_i | N, \beta_k, \gamma_k)}{Z(a_k, b_k, \beta_k, \gamma_k)} - \lambda_t \cdot \text{dist}(\hat{p}_{mle},
     \mathcal{R}_k) \right]$
     \State $\theta \leftarrow \theta - \eta \nabla_{\theta} \mathcal{L}(\theta)$.
\EndFor
\State \textbf{Return:} Trained Meta-Learner $\mathcal{M}_{\theta}$.
\end{algorithmic}
\end{algorithm}

\begin{algorithm}[t]
\caption{Sequential Radius Estimation with Dynamic Fast-Exits}\label{alg:certification}
\begin{algorithmic}[1]
\State \textbf{Input:} Test image $x$, noise level $\sigma$, target classifier $f$, Meta-Learner $\mathcal{M}$.
\State \textbf{Parameters:} Significance $\alpha$, batch size $B$, max samples $N_{max}$, failure threshold $\epsilon_{fail}$, parameters $\epsilon_{start},       \epsilon_{end}$.%
\State \textbf{Phase I: Holdout and Prior Initialization}
\State Extract $\phi(x)$ and clean-image softmax $\mathbf{p}(x)$ from $f$.
\State Draw $N_{sel}$ samples to identify target class $c_A$ and empirical mean $\hat{p}_{sel}$.
\State $\mathcal{M}(\phi(x), \mathbf{p}(x), \hat{p}_{sel}) \to \{ w_k, \beta_k, \gamma_k, \mathcal{R}_k \}_{k=1}^K$.
\State \textbf{Phase II: Sequential Certification}
\State $W_0(p_0) \leftarrow 1$ for all $p_0 \in [0, 1]$; $r_{lcb} \leftarrow 0$; $\mathcal{H} \leftarrow \emptyset$ \Comment{Initialize wealth, radius, and radius history}
\For{$t = 1, 2, \ldots, N_{max}-N_{sel}$}
    \State Compute $W_t(p_0)$ using Mixture of Truncated Betas.
    \If{$t \pmod B == 0$ \textbf{or} $t = N_{max}-N_{sel}$}
        \State $\underline{p_t} \leftarrow \text{BrentSolver}(W_t(p_0) = 1/\alpha)$; $r_{lcb} \leftarrow \sigma \Phi^{-1}(\max(\underline{p_t}, 0.5))$.
        \State $r_{mle} \leftarrow \sigma \Phi^{-1}(\max(\hat{p}_{mle}, 0.5))$; Append $r_{lcb}$ to $\mathcal{H}$.
        \If{$W_t(0.5) \geq 1/\alpha$ and $\hat{p}_{mle} < 0.5$} \textbf{Return} $r = 0$ \Comment{UCB Exit}
        \ElsIf{$W_t(0.5) \leq \epsilon_{fail}$} \textbf{Return} $r = 0$ \Comment{Bankruptcy Exit}
        \Else
            \State $\epsilon_t \leftarrow \Delta \cdot [ \epsilon_{start} - (\epsilon_{start} - \epsilon_{end}) \cdot \frac{t}{N_{max}-N_{sel}} ] \cdot b(r_{mle})$.
            \State \textit{Plateau} $\leftarrow |\mathcal{H}| \ge 4$ \textbf{and} $(\mathcal{H}_{last} - \mathcal{H}_{last-3}) < 0.05 \cdot \mathcal{H}_{last-3}$.
            \If{$(r_{mle} - r_{lcb}) \leq \epsilon_t$ \textbf{or} \textit{Plateau}} \textbf{Return} $r = r_{lcb}$
            \EndIf        
        \EndIf
    \EndIf
\EndFor
\State \textbf{Return} $r = r_{lcb}$.
\end{algorithmic}
\end{algorithm}

\paragraph{Datasets and Backbone Architecture} Our experiments consider attacks against MNIST \citep{lecun1998gradient} (GNU v3.0 license), CIFAR-$10$ \citep{krizhevsky2009learning} (MIT license), and ImageNet \citep{ russakovsky2015imagenet} (which uses a custom, non-commercial license). In the case of models defended by randomised smoothing, each model was trained in PyTorch~\citep{NEURIPS2019_9015}

We trained three backbone models as the base classifiers, each of which were trained to robust under Gaussian noise through augmentations drawn from $\mathcal{N}(0, \sigma^2 I)$. For MNIST and CIFAR-10, we employed a ResNet-18 architecture, which was modified for a single input channel for MNIST. In the case of ImageNet, we considered a ResNet-50 architecture. Features employed by the Meta-Learner are extracted from the final \texttt{avgpool} layer of their respective backbones (512-dimensional for ResNet-18, or 2048-dimensional with a linear projection to 512-dimensional for ResNet-50).

\paragraph{Meta-Learner Architecture} The Meta model $\mathcal{M}_\theta$ builds upon a \texttt{LayerNorm} stage followed by a linear projection to 512 dimensions. This is then followed by a sequential MLP mapping, containing two linear layers (512 units) with \texttt{LayerNorm} and \texttt{ReLU} activations, before branching into independent linear heads for mixture weights ($\pi$), Beta parameters ($\beta, \gamma$), and optionally, dynamic support boundaries ($a, b$). %

\paragraph{Optimization and Training} As discussed in the main body of the text, the meta-learner is trained to maximize the \textbf{Expected Log-Wealth} (Kelly Criterion) over full bitstream sequences. Optimization employed an Adam learner for $60$ epochs with a fixed learning rate of $10^{-3}$ and a weight decay of $10^{-4}$. To support training, we employed randomized deterministic data splits, with $5,000$ samples employed for training with MNIST and CIFAR-10, and $10,000$ samples for ImageNet. In all cases, the amount of evaluation samples was set to $10\%$ of the training samples. The base models for MNIST and CIFAR-10 were trained on a 12GB RTX2080 Ti GPU, with a training time of less than $1$ hour per $\sigma$. For speed, ImageNet was trained on a 80GB H100 GPU, with a total training time of $2$ hours per $\sigma$. Meta-learner training was performed on the 2080Ti, taking about $2$ minutes per each Meta-learner for ImageNet, and less than $90$ seconds for MNIST and CIFAR-10. 

\paragraph{Differentiable Truncation via Series Expansion} Training the meta-learner with a Kelly-optimal loss on truncated Beta distributions requires a differentiable implementation of the regularized incomplete Beta       function, $I_x(a, b)$. As implementations of $I_x$ are typically not compatible with automatic differentiation, we employ a $4^{th}$-order Taylor series expansion to approximate the integral mass in the cases where derivatives are required. For a given support boundary $x \in [a_k, b_k]$, the log-mass is approximated as
\begin{equation}
    \log I_x(\alpha, \beta) \approx a \log x + b \log(1-x) - \log a - \log B(a, b) + \log \left( 1 + \sum_{n=1}^{4} T_n \right)\enspace,
\end{equation}
where $T_n$ corresponds to the Pochhammer ratios
\begin{equation}
    T_1 = \frac{a+b}{a+1}x, \quad T_n = T_{n-1} \cdot \frac{a+b+n-1}{a+n}x\enspace.
\end{equation}
To ensure numerical stability across the entire $[0, 1]$ probability range, we leverage the symmetry property $I_x(a, b) = 1 - I_{1-x}(b, a)$. When the truncation boundary is $x > 0.5$, the model computes the mass in the complement space using the same expansion, preventing the vanishing gradient issues associated with the high-probability tails of the Beta distribution.

\paragraph{Certification} For certifications, we set the error probability $\alpha$ to $0.001$, corresponding to a $99.9\%$ confidence interval. Unless otherwise stated, the selection glimpse took place over $N_{sel} = 100$, and similarly the checking interval $B$ was also set to $100$. All certifications were capped at $10,000$ max samples.

\subsection{Radius-Specialized Stopping Dynamics}\label{app:radius-specialized}

To articulate the influence of focusing computational allocation across varying difficulty levels, we define two specialists---\textbf{Small-R} and \textbf{Large-R}---using a biased stopping criterion. These conditions incorporate an augmented sequential exit condition defined by the precision threshold $\epsilon_t$, whereby 
\begin{equation}
     (R_{mle} - R_{lcb}) \le \epsilon_t \cdot b(R_{mle})\enspace.
\end{equation}
Here $R_{mle}$ is the current estimated radius, $R_{lcb}$ is the anytime-valid lower bound, and $b(R_{mle})$ is a radius-dependent bias factor.

For the Small-R specialist, we optimize for $R < 0.5$. This specialist employs an aggressive bias $b = 0.2$ when $R_{mle} < 0.5$, facilitating rapid exit for low-margin samples. The Large-R specialist is optimized for $R > 1.0$ by way of a bias $b = 0.6$ when $R_{mle} > 1.0$. 

For samples falling outside the radius regime, the bias $b$ is increased to $2.0$, to force the specialist to either exit immediately (if the plateua condition is met), or to continue sampling until high precision is achieved. In doing so, the specialists effectively de-prioritize non-target compute, allowing us to demonstrate that specialization is a tool that be employed to leverage the anytime-valid properties of E-values to dynamically adjust sample budgets without sacrificing statistical integrity. 

\subsection{Computational Costs}

A critical consideration for sequential certification frameworks is the wall-clock overhead introduced by the decision-making logic (meta-learning and interval checks) relative to the time saved by reducing base model forward passes. In this section, we provide a holistic breakdown of the computational costs associated with the Meta-RS framework. At inference time, the Meta-RS model involves \textbf{Feature Extraction} from the penultimate layer embedding from the backbone model (conducted at $N=100$); and \textbf{Meta-Learner Inference} as a single forward pass through the 4-layer Residual MLP to predict the image-specific prior. After the Meta-RS model has been produced the prediction of the prior, \textbf{Brent-Dekker Root Finding} is performed every $B$ steps, which is required to search for the LCB. 

We benchmarked these components on an NVIDIA 2080Ti GPU using ResNet-18 (CIFAR-10) and ResNet-50 (ImageNet) backbones. Table~\ref{tab:wall_clock} summarizes the results.

\begin{table}[h]
\centering
\small
\setlength{\tabcolsep}{8pt}
\caption{\textbf{Wall-Clock Timing Breakdown}. We compare the marginal overhead of the Meta-RS components against the cost of base model forward passes on a 2080Ti GPU.}
\label{tab:wall_clock}
\begin{tabular}{lcc}
\toprule
\textbf{Action} & \textbf{Component} & \textbf{Latency (ms)} \\
\midrule
\multirow{2}{*}{\textbf{Backbone Cost}} & ResNet-18 (100-sample batch) & 3.8 \\
& ResNet-50 (100-sample batch) & 102.9 \\
\midrule
\multirow{2}{*}{\textbf{Meta-RS Overhead}} & Meta-Learner & 0.9 \\
& Brent-Dekker Search (Per Check) & 1.9 \\
\bottomrule
\end{tabular}
\end{table}

The total computational overhead for a full $N=10,000$ ImageNet certification is \textbf{10.4 s} (comprising 1 batch for class prediction and 100 batches). In contrast, if the Meta learner is able to produce a $10$ fold reduction in the number of samples, then the total cost is \textbf{1.15 s}, an $89\%$ reduction in the computational cost for ImageNet. This confirms that the computational cost of the anytime-valid logic is negligible—accounting for less than $2\%$ of the total budget—allowing the sample complexity gains to translate directly into massive operational speedups in high-optimized inference systems.

\section{Champion Selection Processes}\label{app:champion_selection}

To evaluate the operational versatility of our framework, we identify three distinct \emph{Champion} archetypes. Each represents a specific optimization of our Meta-learner architecture designed to address different real-world deployment constraints. In short, the Global Champion demonstrates generalisability of the meta-learner, and provides a universal speedup for standard certification; while the Specialists demonstrate the framework’s ability to maximize performance.

Of these, the Global Champion was set as \texttt{Meta-1-Margin-Hybrid} across all experiments. This technique produced the optima of 
\begin{equation}
    \min_{\text{Architecture}} \bar{N} \quad \text{subject to} \quad \text{Acc}_{\text{Meta}} \ge \text{Acc}_{\text{Cohen, 10k}} - 0.02
\end{equation}
when testing across all datasets and noise levels, based upon a sweep over the architectural space ($K \in \{1, 3, 6, 10\}$, Entropy or Margin feature modes, and ranges). This configuration demonstrates that a single meta-learned prior can capture universal statistical convictions across multiple semantic domains. 

For each dataset, we constructed an Accuracy Champion (otherwise labeled as the Dataset Specialist), which restricted the above optimization criteria to a single dataset-and-noise configuration. The choice of champion for each dataset can be found in Table~\ref{tab:champion_definitions}.

Finally, the Efficiency Champion is optimized using a similar objective, targeting parity with a moderate Cohen-1k baseline
\begin{equation}
    \min_{\text{Architecture}} \bar{N} \quad \text{subject to} \quad \text{Acc}_{\text{Meta}} \ge \text{Acc}_{\text{Cohen, 1k}}\enspace.
\end{equation}

\begin{table}[h]
\centering
\small
\setlength{\tabcolsep}{4pt}
\caption{\textbf{Champion Configurations by Dataset and Noise Level}. We define three champion archetypes: \textbf{Global} (fixed Meta-1-margin-hybrid), \textbf{Efficiency} (absolute minimum average sample count $\bar{N}$), and \textbf{Accuracy} (minimum $\bar{N}$ maintaining accuracy parity with Cohen-10k). All configurations use an aggression factor $\lambda = 1.2$.}
\label{tab:champion_definitions}
\begin{tabular}{lcccc}
\toprule
Dataset & Sigma ($\sigma$) & Global Champion & Efficiency Champion & Accuracy Champion \\
\midrule
\multirow{3}{*}{CIFAR-10}  & 0.25 & Meta-1-Margin-Hybrid & Meta-3-Entropy-Dynamic & Meta-1-Entropy-Dynamic \\
  & 0.5 & Meta-1-Margin-Hybrid & Meta-1-Entropy-Dynamic & Meta-1-Entropy-Dynamic \\
  & 1. & Meta-1-Margin-Hybrid & Meta-6-Entropy-Dynamic & Meta-3-Entropy-Dynamic \\
  \midrule
  \multirow{3}{*}{ImageNet} & 0.25 & Meta-1-Margin-Hybrid & Meta-3-Margin-Hybrid & Meta-6-Entropy-Hybrid \\
  & 0.5 & Meta-1-Margin-Hybrid & Meta-1-Entropy-Hybrid & Meta-1-Entropy-Hybrid \\
  & 1. & Meta-1-Margin-Hybrid & Meta-3-Margin-Dynamic & Meta-3-Margin-Dynamic \\
  \midrule
\multirow{3}{*}{MNIST}  & 0.25 & Meta-1-Margin-Hybrid & Meta-1-Margin-Dynamic & Meta-1-Margin-Dynamic \\
  & 0.5 & Meta-1-Margin-Hybrid & Meta-1-Entropy-Hybrid & Meta-1-Entropy-Hybrid \\
  & 1. & Meta-1-Margin-Hybrid & Meta-1-Margin-Dynamic & Meta-1-Margin-Dynamic \\
\bottomrule
\end{tabular}
\end{table}

\section{Detailed Results and Aggregated Performance}

\begin{figure}[h]
    \centering
    \includegraphics[width=\textwidth]{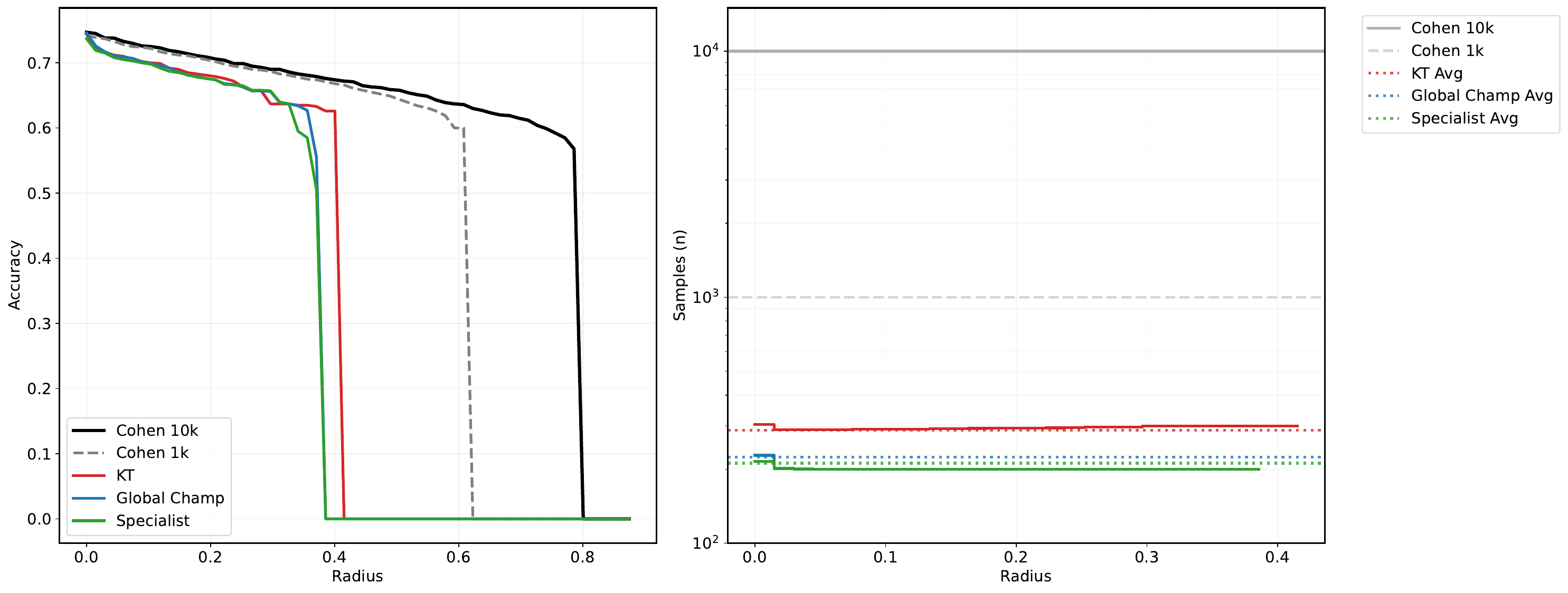}
    \caption{Certification and efficiency results for ImageNet at $\sigma=0.25$. Curves compare the Anytime-Valid Global Champ and Specialist Meta-models against Cohen-10k/1k and KT Prior baselines.}
    \label{fig:results_imagenet_s025} 
\end{figure}

\begin{figure}[h]
    \centering
    \includegraphics[width=\textwidth]{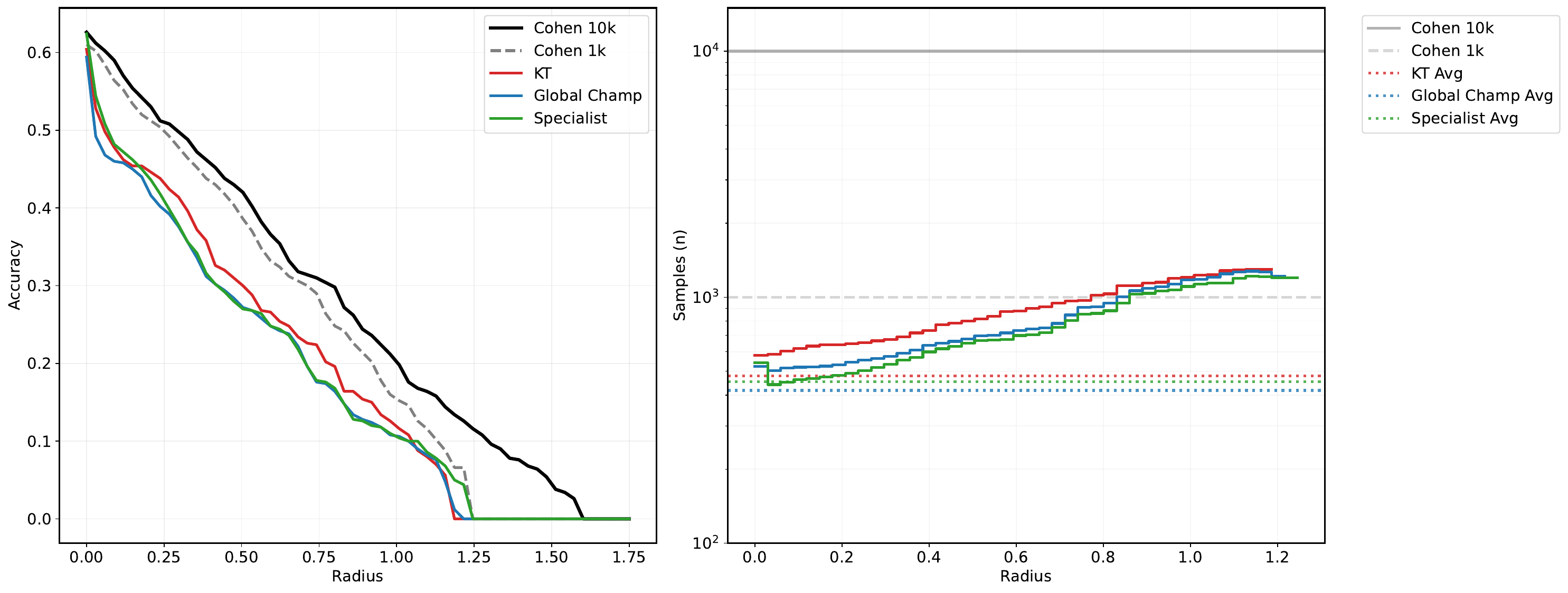}
    \caption{Certification and efficiency results for CIFAR10 at $\sigma=0.5$. Curves compare the Anytime-Valid Global Champ and Specialist Meta-models against Cohen-10k/1k and KT Prior baselines.}
    \label{fig:results_cifar10_s05}
\end{figure}

The full suite of certification and efficiency results for individual datasets and noise levels is provided in Figures~\ref{fig:results_imagenet_s025} through to \ref{fig:results_cifar10_s05}. These figures are included to demonstrate the framework's consistent performance across three distinct data manifolds (MNIST, CIFAR-10, ImageNet) and various noise levels. By visualizing the \emph{Average Sample Count} (dashed lines) alongside the \emph{Certified Accuracy} (solid lines), we provide empirical proof that the meta-learned priors effectively deliver 20x--45x speedups over \citet{cohen2019certified} while maintaining a similar accuracy profile. When comparing against the KT prior, our Meta-learner is able to consistently reduce the required number of samples to both certify and reject samples. This serves as the primary evidence for the system's operational viability in high-throughput environments.

\begin{table}
\small
\centering
\caption{\textbf{Comprehensive performance comparison across datasets and noise levels.}}
\label{tab:full_results}
\begin{tabular}{lllrcccc}
\toprule
Dataset & Sigma & Strategy & Acc (\%) & Mean Rad & Avg All & Avg Cert & Avg Rej \\
\midrule
\multirow{15}{*}{ImageNet}  & 0.25 & KT & 72.6\% & 0.398 & 934.7 & 1211.7 & 200.7 \\
  & 0.25 & Global Champ & 72.4\% & 0.412 & 855.5 & 1105.4 & 200.0 \\
  & 0.25 & Specialist & 72.3\% & 0.412 & 855.3 & 1106.2 & 200.4 \\
  & 0.25 & Cohen-10k & 74.7\% & 0.536 & 10000 & 10000 & 10000 \\
  & 0.25 & Cohen-1k & 74.2\% & 0.420 & 1000 & 1000 & 1000 \\
  & 0.5 & KT & 71.3\% & 0.748 & 940.3 & 1237.4 & 202.1 \\
  & 0.5 & Global Champ & 71.2\% & 0.716 & 1007.2 & 1333.6 & 200.3 \\
  & 0.5 & Specialist & 71.2\% & 0.716 & 1000.9 & 1324.7 & 200.3 \\
  & 0.5 & Cohen-10k & 72.6\% & 0.997 & 10000 & 10000 & 10000 \\
  & 0.5 & Cohen-1k & 72.0\% & 0.787 & 1000 & 1000 & 1000 \\
  & 1. & KT & 68.4\% & 1.364 & 901.3 & 1221.8 & 207.6 \\
  & 1. & Global Champ & 68.5\% & 1.410 & 851.3 & 1149.6 & 202.5 \\
  & 1. & Specialist & 68.5\% & 1.387 & 844.3 & 1139.4 & 202.5 \\
  & 1. & Cohen-10k & 69.7\% & 1.801 & 10000 & 10000 & 10000 \\
  & 1. & Cohen-1k & 68.9\% & 1.440 & 1000 & 1000 & 1000 \\
  \midrule
  \multirow{15}{*}{MNIST} & 0.25 & KT & 99.2\% & 0.573 & 1264.8 & 1273.4 & 200.0 \\
  & 0.25 & Global Champ & 99.2\% & 0.599 & 1163.6 & 1171.4 & 200.0 \\
  & 0.25 & Specialist & 99.2\% & 0.585 & 1162.2 & 1170.0 & 200.0 \\
  & 0.25 & Cohen-10k & 99.2\% & 0.773 & 10000 & 10000 & 10000 \\
  & 0.25 & Cohen-1k & 99.2\% & 0.600 & 1000 & 1000 & 1000 \\
  & 0.5 & KT & 98.4\% & 1.062 & 1199.0 & 1215.0 & 212.5 \\
  & 0.5 & Global Champ & 98.4\% & 1.095 & 1103.0 & 1117.5 & 212.5 \\
  & 0.5 & Specialist & 98.4\% & 1.095 & 1103.0 & 1117.5 & 212.5 \\
  & 0.5 & Cohen-10k & 98.8\% & 1.388 & 10000 & 10000 & 10000 \\
  & 0.5 & Cohen-1k & 98.8\% & 1.123 & 1000 & 1000 & 1000 \\
  & 1. & KT & 89.2\% & 0.981 & 967.8 & 1058.7 & 216.7 \\
  & 1. & Global Champ & 88.6\% & 0.946 & 1059.0 & 1168.8 & 205.3 \\
  & 1. & Specialist & 89.2\% & 0.993 & 925.8 & 1011.7 & 216.7 \\
  & 1. & Cohen-10k & 91.2\% & 1.200 & 10000 & 10000 & 10000 \\
  & 1. & Cohen-1k & 89.6\% & 1.070 & 1000 & 1000 & 1000 \\
  \midrule
\multirow{15}{*}{CIFAR-10}& 0.25 & KT & 73.6\% & 0.308 & 741.0 & 934.5 & 201.5 \\
  & 0.25 & Global Champ & 72.2\% & 0.309 & 659.0 & 835.5 & 200.7 \\
  & 0.25 & Specialist & 74.0\% & 0.310 & 678.6 & 846.8 & 200.0 \\
  & 0.25 & Cohen-10k & 78.8\% & 0.407 & 10000 & 10000 & 10000 \\
  & 0.25 & Cohen-1k & 77.8\% & 0.338 & 1000 & 1000 & 1000 \\
  & 0.5 & KT & 58.6\% & 0.377 & 703.2 & 1054.9 & 205.3 \\
  & 0.5 & Global Champ & 57.0\% & 0.376 & 695.8 & 1058.9 & 214.4 \\
  & 0.5 & Specialist & 58.6\% & 0.373 & 665.8 & 991.1 & 205.3 \\
  & 0.5 & Cohen-10k & 62.6\% & 0.473 & 10000 & 10000 & 10000 \\
  & 0.5 & Cohen-1k & 61.0\% & 0.411 & 1000 & 1000 & 1000 \\
  & 1. & KT & 41.8\% & 0.390 & 603.8 & 1152.2 & 210.0 \\
  & 1. & Global Champ & 41.2\% & 0.377 & 665.8 & 1314.6 & 211.2 \\
  & 1. & Specialist & 41.8\% & 0.395 & 573.8 & 1080.4 & 210.0 \\
  & 1. & Cohen-10k & 44.4\% & 0.491 & 10000 & 10000 & 10000 \\
  & 1. & Cohen-1k & 43.8\% & 0.431 & 1000 & 1000 & 1000 \\
\bottomrule
\end{tabular}
\end{table}

\begin{figure}[h]
    \centering
    \includegraphics[width=\textwidth]{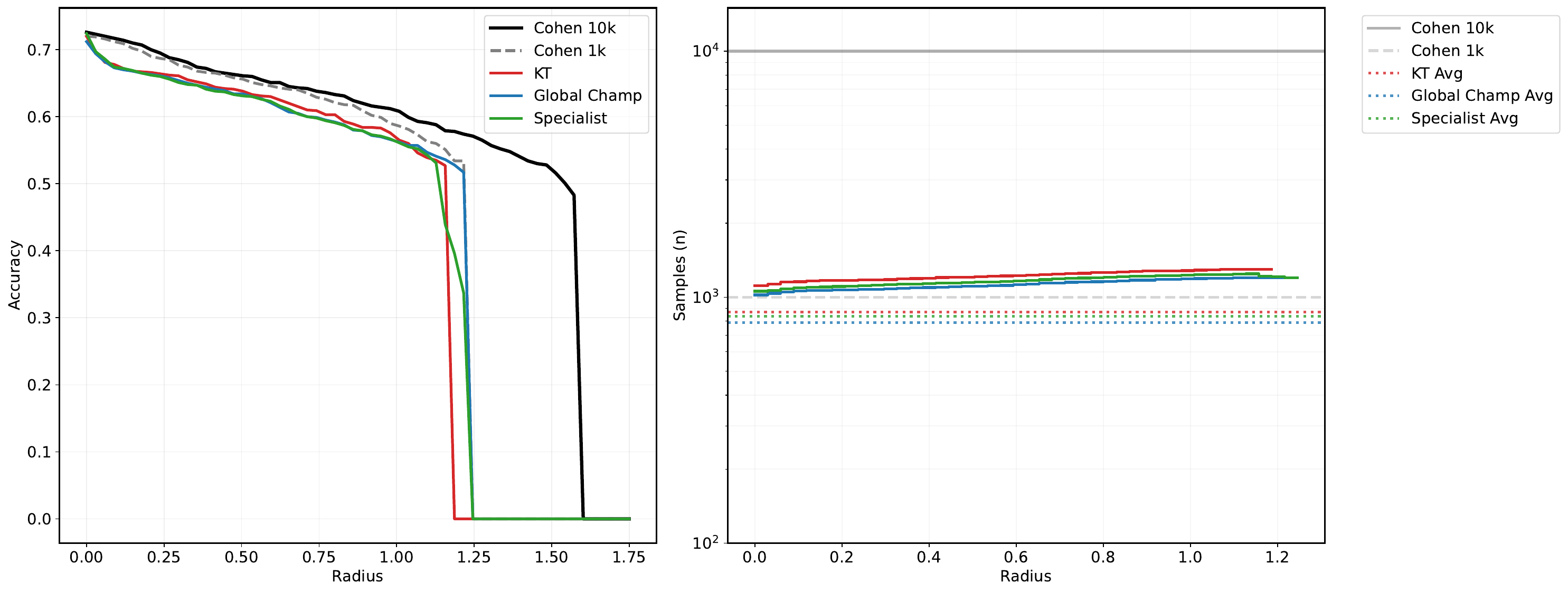}
    \caption{Certification and efficiency results for ImageNet at $\sigma=0.5$. Curves compare the Anytime-Valid Global Champ and Specialist Meta-models against Cohen-10k/1k and KT Prior baselines.}
\end{figure}

\begin{figure}[h]
    \centering
    \includegraphics[width=\textwidth]{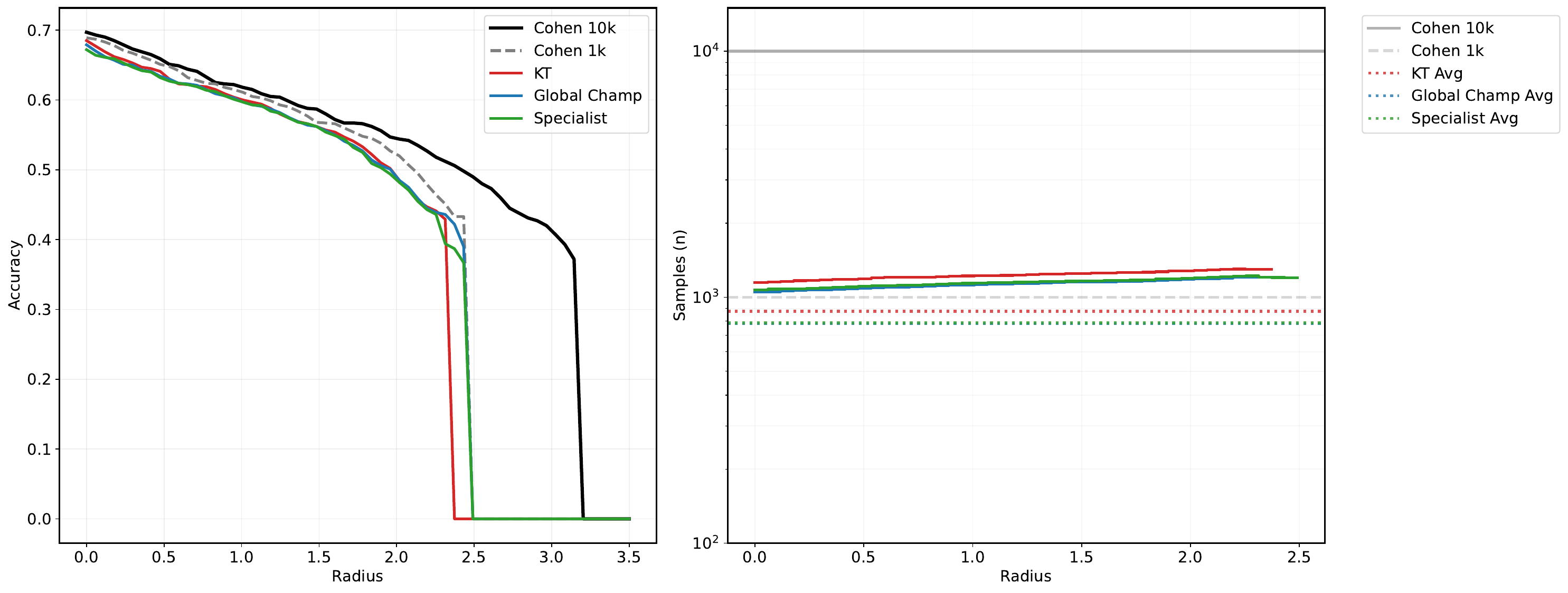}
    \caption{Certification and efficiency results for ImageNet at $\sigma=1.0$. Curves compare the Anytime-Valid Global Champ and Specialist Meta-models against Cohen-10k/1k and KT Prior baselines.}
\end{figure}

\begin{figure}[h]
    \centering
    \includegraphics[width=\textwidth]{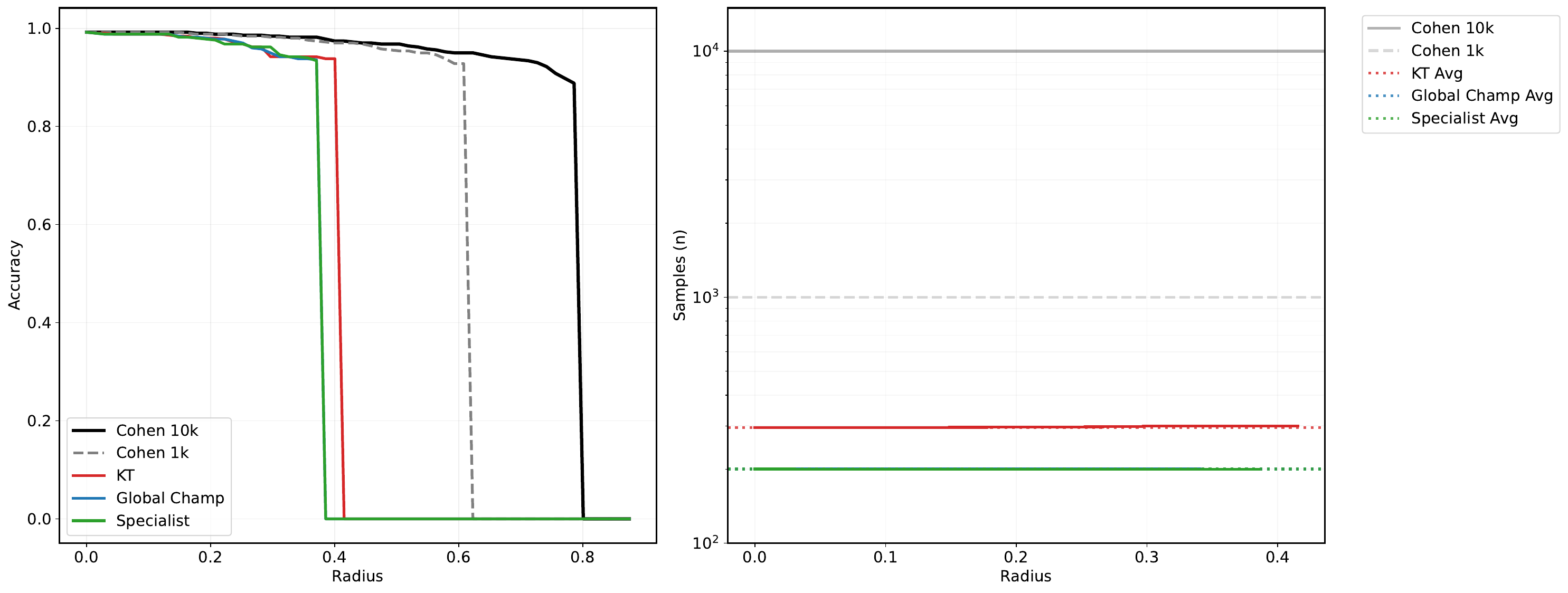}
    \caption{Certification and efficiency results for MNIST at $\sigma=0.25$. Curves compare the Anytime-Valid Global Champ and Specialist Meta-models against Cohen-10k/1k and KT Prior baselines.}
\end{figure}

\begin{figure}[h]
    \centering
    \includegraphics[width=\textwidth]{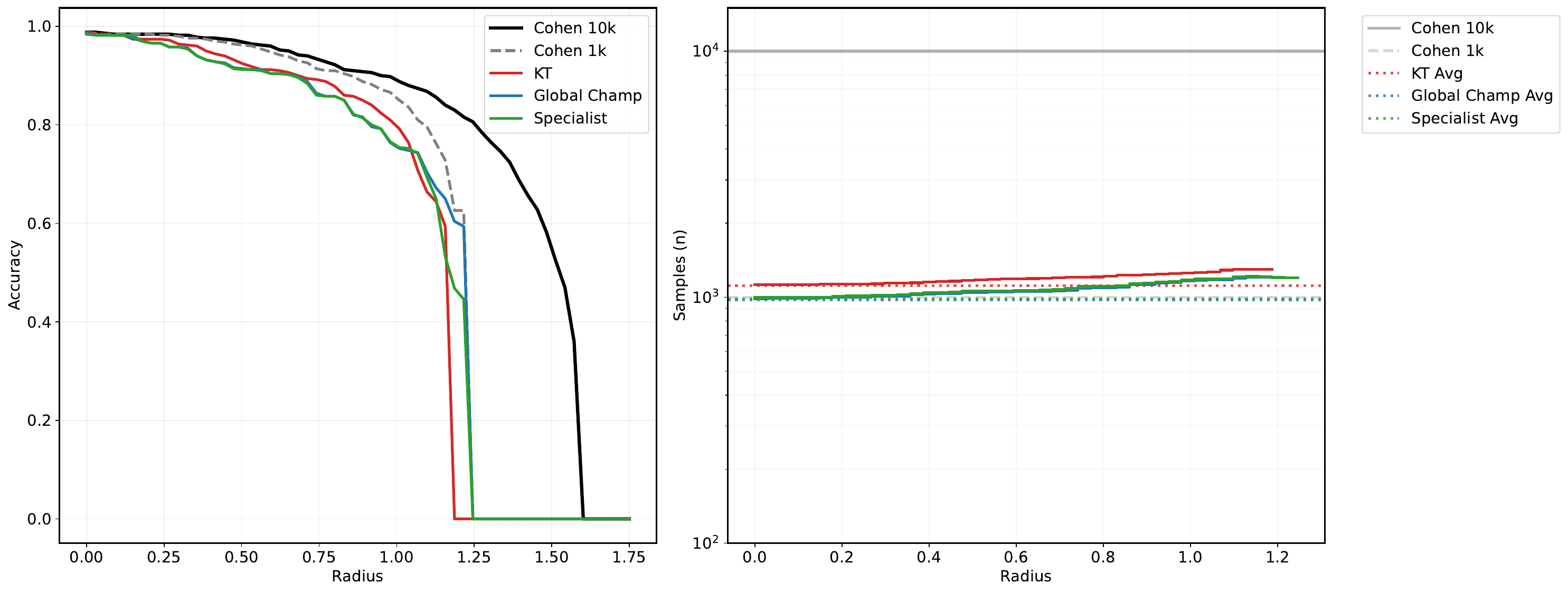}
    \caption{Certification and efficiency results for MNIST at $\sigma=0.5$. Curves compare the Anytime-Valid Global Champ and Specialist Meta-models against Cohen-10k/1k and KT Prior baselines.}
\end{figure}

\begin{figure}[h]
    \centering
    \includegraphics[width=\textwidth]{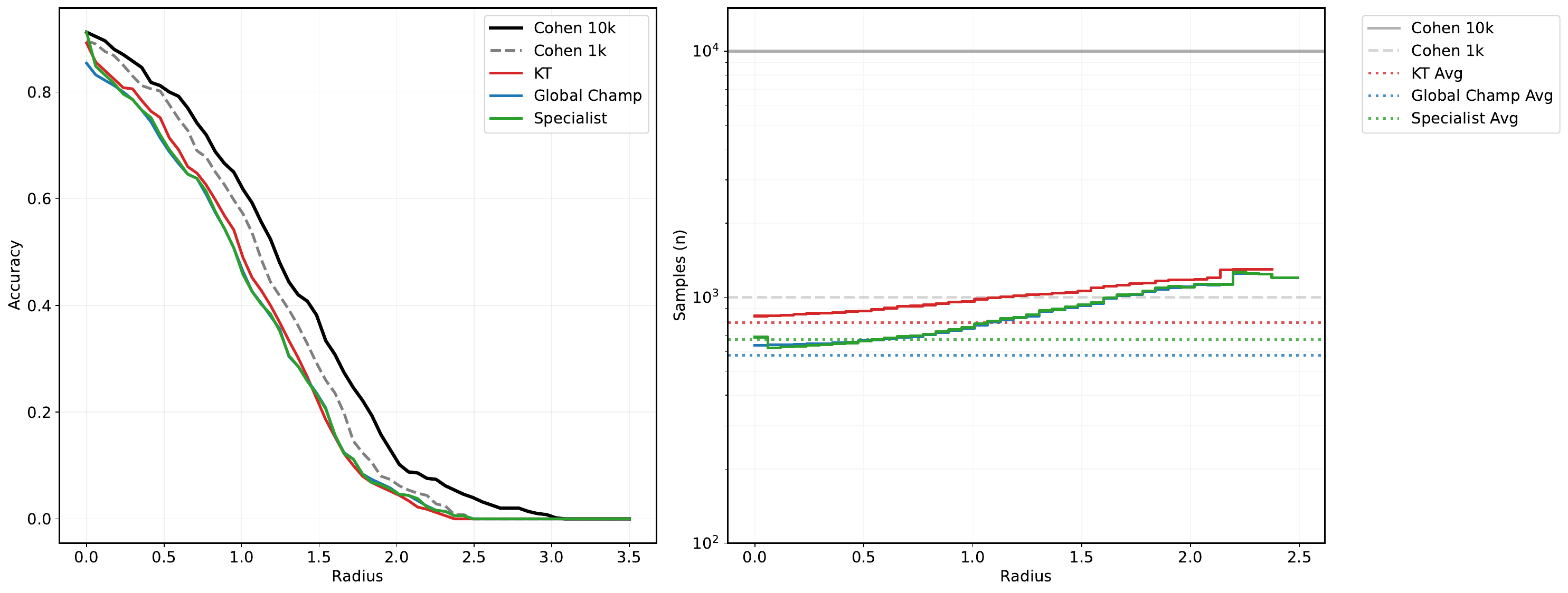}
    \caption{Certification and efficiency results for MNIST at $\sigma=1.0$. Curves compare the Anytime-Valid Global Champ and Specialist Meta-models against Cohen-10k/1k and KT Prior baselines.}
\end{figure}

\begin{figure}[h]
    \centering
    \includegraphics[width=\textwidth]{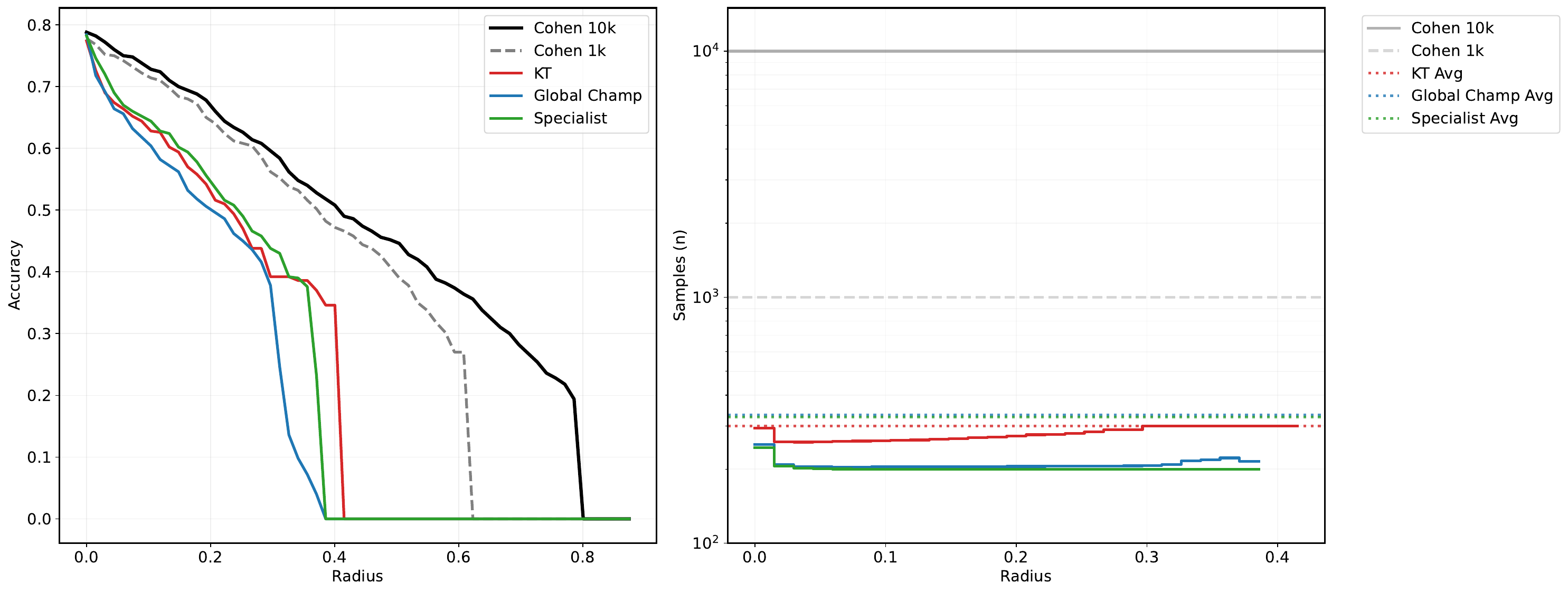}
    \caption{Certification and efficiency results for CIFAR10 at $\sigma=0.25$. Curves compare the Anytime-Valid Global Champ and Specialist Meta-models against Cohen-10k/1k and KT Prior baselines.}
\end{figure}

\section{Ablation Studies and Sensitivity Analysis}

This section provides empirical justification for the architectural and heuristic choices in the Meta-RS framework. All studies are conducted using ImageNet at $\sigma=0.25$ unless otherwise specified.

\begin{figure}[h]
    \centering
    \includegraphics[width=0.7\textwidth]{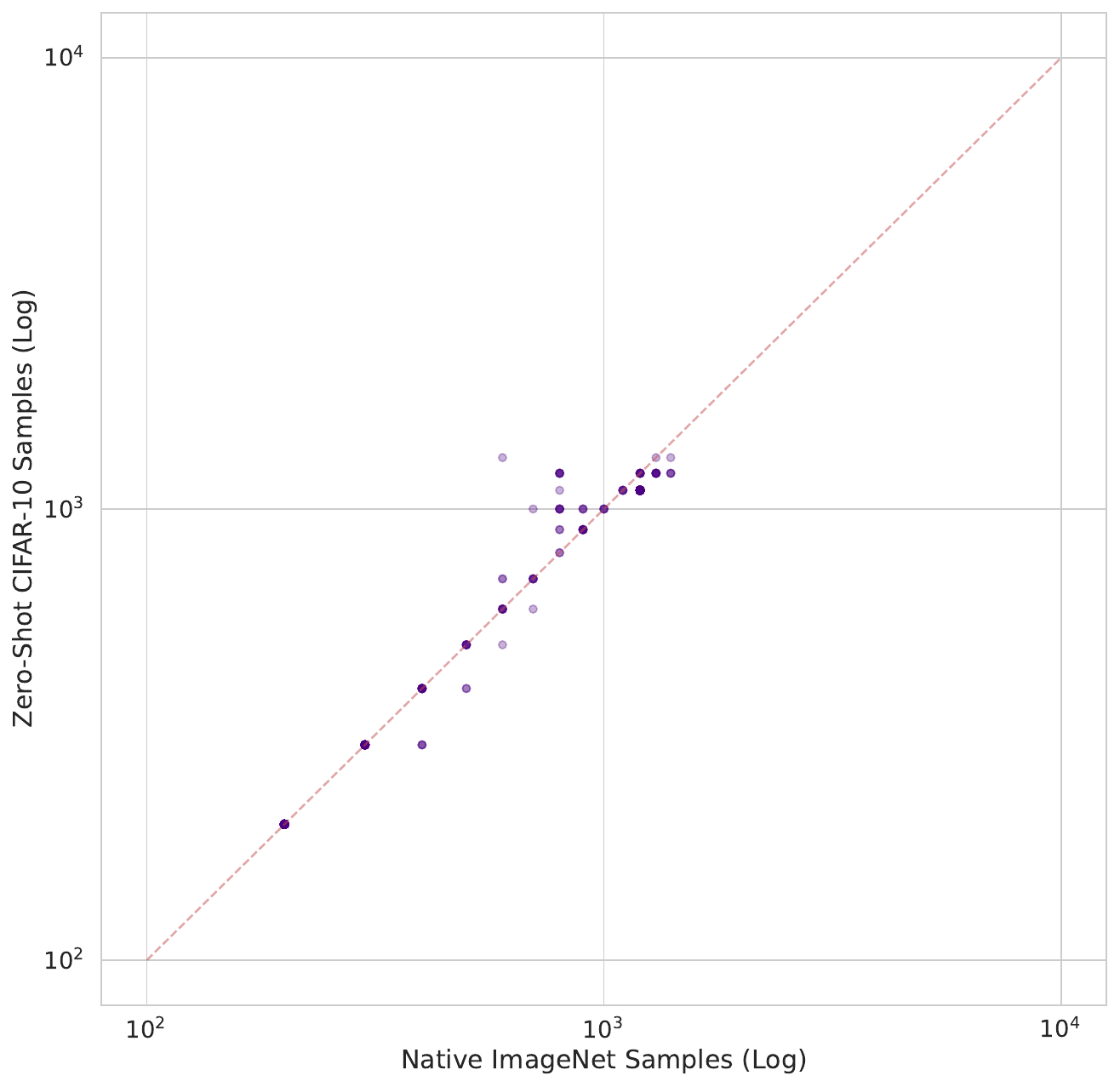}
    \caption{\textbf{Zero-Shot Transferability.} A log-log scatter plot correlating termination latencies of an ImageNet-native prior vs. a CIFAR-trained prior tested on ImageNet. These results suggest that the meta-learner captures universal statistical properties of classifier conviction that transcend specific datasets.}
    \label{fig:ablation_4}
\end{figure}

\paragraph{Statistical Universality and Zero-Shot Transferability} (Figure~\ref{fig:ablation_4}). This log-log scatter plot correlates termination latencies across datasets, and demonstrates that the meta-learner may be able to capture fundamental statistical properties of classifier conviction rather than dataset-specific artifacts, justifying its use as a plug-and-play engine for new models.

\begin{figure}[h]
    \centering
    \includegraphics[width=0.48\textwidth]{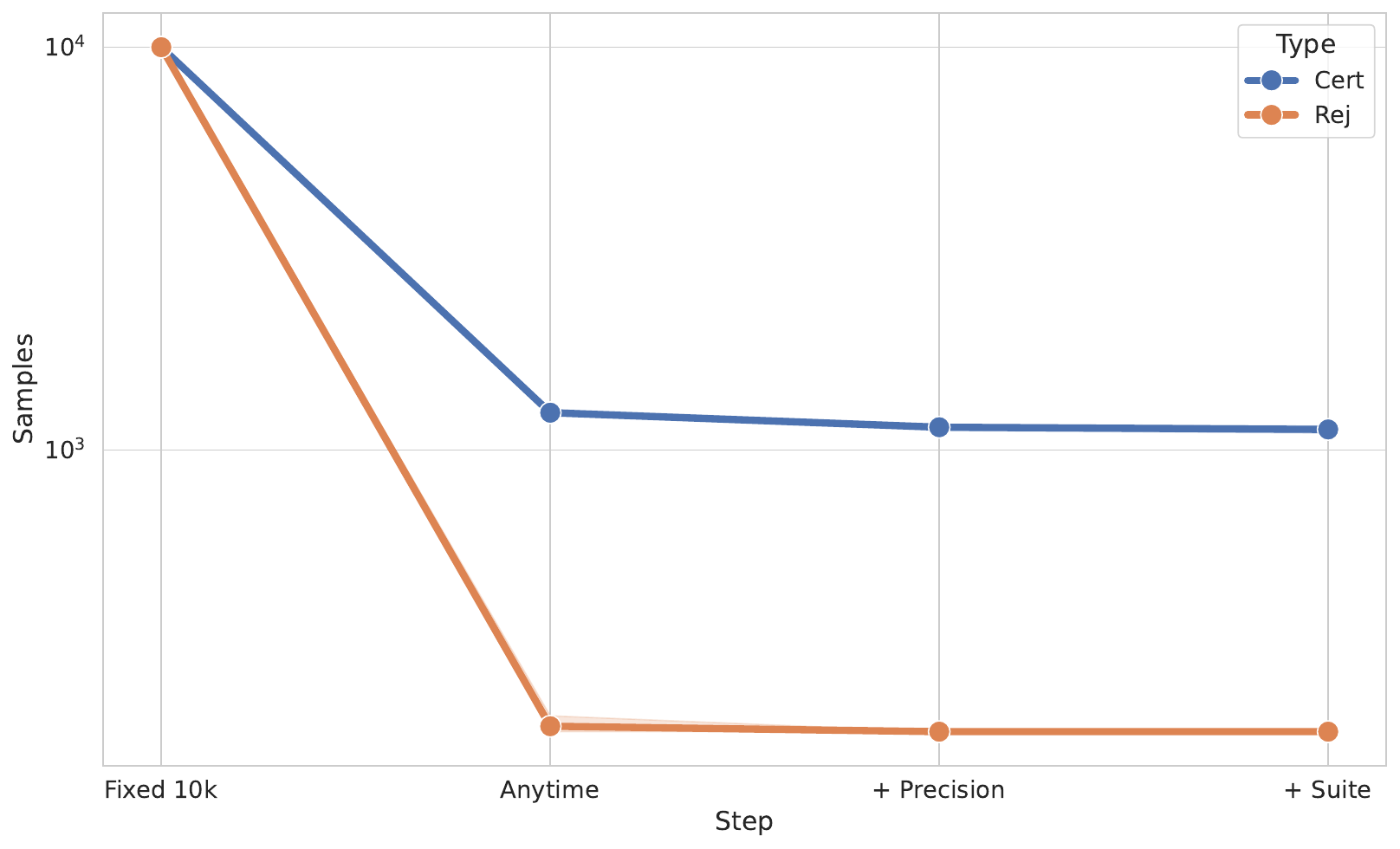}
    \caption{\textbf{Algorithmic Logic Waterfall.} Incremental gains from Anytime-Valid LCB, Precision Stopping, and Rejection Heuristics. The comparison shows that while Precision Stopping provides the bulk of certification speedup, the Bankruptcy/UCB rejection suite is critical for minimizing the cost of non-robust samples.}
    \label{fig:ablation_5}
\end{figure}

\paragraph{Algorithmic Logic Waterfall} Figure~\ref{fig:ablation_5} breaks down the efficiency of our approach into the drivers. It is included to demonstrate the additive value of our Dual-Exit strategy, showing that while precision stopping drives certification, with performance being further refined by the Bankruptcy/UCB suite.

\begin{figure}[h]
    \centering
    \includegraphics[width=0.7\textwidth]{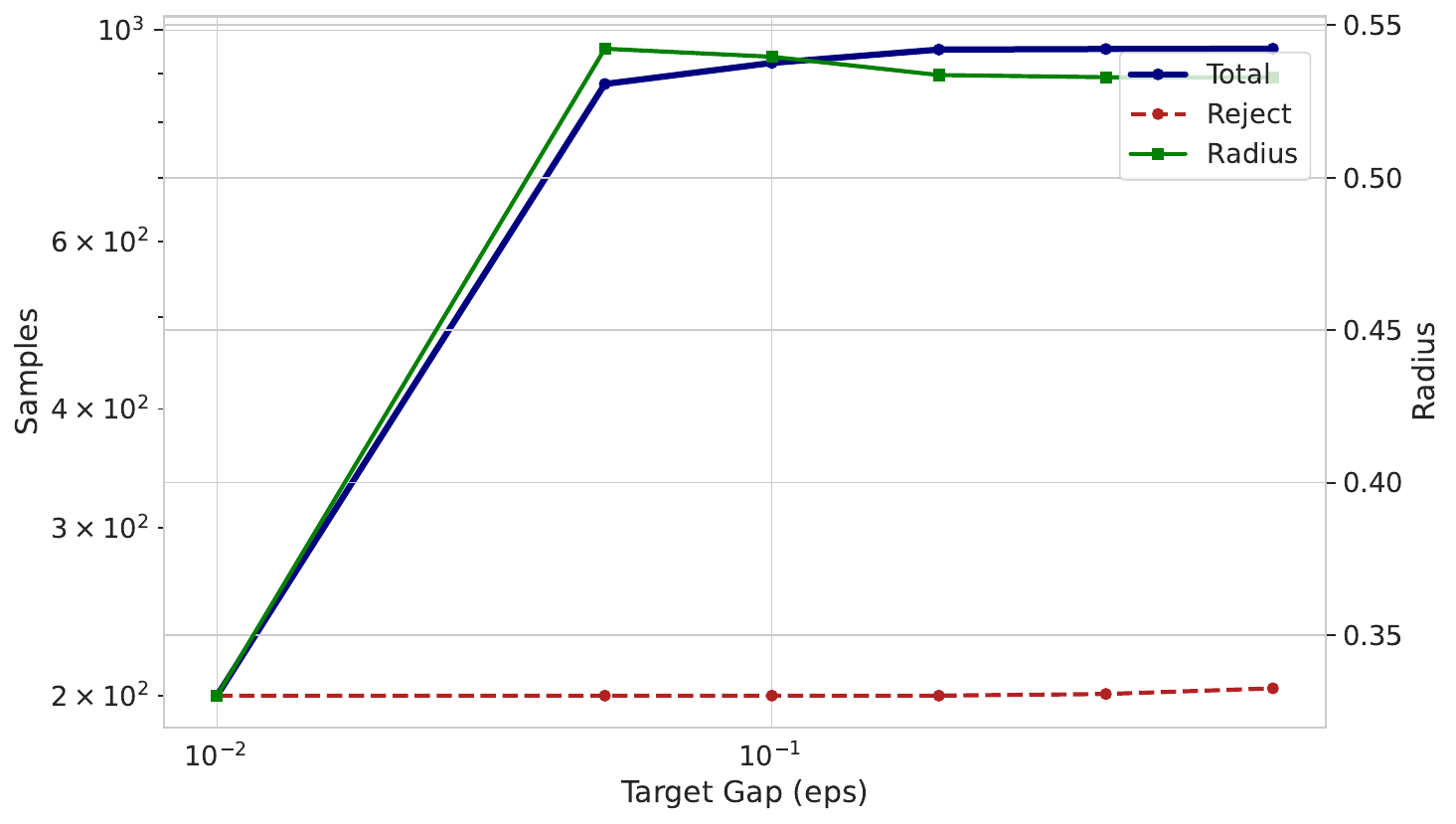}
    \caption{\textbf{Precision-Efficiency Pareto Frontier.} This plot maps the scaling of sample complexity against the radius precision target $\epsilon$. As the requirement for tightness increases ($\epsilon \to 0.01$), the sample cost grows logarithmically, allowing practitioners to choose an operating point based on their computational budget.}
    \label{fig:ablation_7}
\end{figure}

\paragraph{Precision-Efficiency Pareto Frontier} (Figure~\ref{fig:ablation_7}). This plot maps sample complexity against radius precision target $\epsilon$, confirming the influence of this parameterization upon resource demands.

\subsection{Parameter Influence}

To interrogate the influence of different parameter strategies, we considered meta-model variants across slices  in terms of the range (Table~\ref{tab:range_strategies}), mixture component counts (Table~\ref{tab:mixture_strategies}), and input features (Table~\ref{tab:input_strategies}). The clear signal from these is the \emph{lack of signal}, in that on average there is no consistent trends to any of these approaches. This is not to say that these factors are not influential, but rather that the drivers of performance are multifactorial. 

\begin{table}[h]
\centering
\small
\setlength{\tabcolsep}{4pt}
\caption{\textbf{Global Influence of prior range strategies}. Average influence of different range settings for the Meta-model, radii relative to \citet{cohen2019certified}.}
\label{tab:range_strategies}
\begin{tabular}{lrr}
\toprule
Range & Avg Cert Samples & Mean Radius (rel. Cohen) \\
\midrule
Dynamic & 1071.631 & 0.773 \\
Full & 1254.115 & 0.731 \\
Hybrid & 1123.796 & 0.781 \\
\bottomrule
\end{tabular}
\end{table}

\begin{table}[h]
\centering
\small
\setlength{\tabcolsep}{4pt}
\caption{\textbf{Global Influence of mixture components (K)}. Average influence of different range settings for the Meta-model, radii relative to \citet{cohen2019certified}.}
\label{tab:mixture_strategies}
\begin{tabular}{lrr}
\toprule
 K & Avg Cert Samples & Mean Radius (rel. Cohen) \\
\midrule
     1 & 1130.593 & 0.761 \\
     3 & 1153.369 & 0.761 \\
     6 & 1156.495 & 0.763 \\
    10 & 1158.932 & 0.762 \\
    \bottomrule
\end{tabular}
\end{table}

\begin{table}[h]
\centering
\small
\setlength{\tabcolsep}{4pt}
\caption{\textbf{Global Influence of input features}. Average influence of different range settings for the Meta-model, radii relative to \citet{cohen2019certified}.}
\label{tab:input_strategies}
 \begin{tabular}{lrr}
 \toprule
 Input Features & Avg Cert Samples & Mean Radius (rel. Cohen) \\  
 \midrule
 entropy & 1148.657 & 0.762 \\
 margin & 1151.037 & 0.761 \\
 \bottomrule
 \end{tabular}
 \end{table}

\end{document}